\documentclass[journal]{IEEEtran}

\usepackage{amsmath,amssymb,amsfonts,mathtools}
\usepackage{graphicx}
\usepackage{booktabs}
\usepackage{cite}
\usepackage{url}
\usepackage[hidelinks]{hyperref}
\usepackage{newtxtext,newtxmath}
\usepackage{algorithm}
\usepackage{algpseudocode}
\usepackage{float}
\usepackage{placeins}
\usepackage{subcaption}
\usepackage{makecell}
\usepackage{array}
\usepackage{multirow}
\usepackage{adjustbox}
\usepackage{siunitx}
\usepackage{tabularx}
\usepackage[dvipsnames]{xcolor}

\graphicspath{{Images/}}

\renewcommand{\arraystretch}{1.12}
\newcolumntype{Y}{>{\centering\arraybackslash}X}

\definecolor{darkgreen}{RGB}{0,120,60}
\definecolor{darkred}{RGB}{150,0,0}
\definecolor{darkblue}{RGB}{0,60,140}

\newcommand{\E}{\mathbb{E}}
\newcommand{\Pp}{\mathbb{P}}

\newcommand{\cX}{\mathcal{X}}

\newcommand{\cZ}{\mathcal{Z}}
\newcommand{\cU}{\mathcal{U}}
\newcommand{\cM}{\mathcal{M}}
\newcommand{\cP}{\mathcal{P}}
\newcommand{\cG}{\mathcal{G}}
\newcommand{\cA}{\mathcal{A}}
\newcommand{\cI}{\mathcal{I}}
\newcommand{\cD}{\mathcal{D}}

\newcommand{\safeincludegraphics}[2][]{%
  \IfFileExists{#2}{\includegraphics[#1]{#2}}{%
    \fbox{\parbox{0.92\linewidth}{\centering Missing figure file: \texttt{\detokenize{#2}}}}%
  }%
}

\title{UCATSC: Uncertainty-Aware Constrained Traffic Signal Control Under Vision-Based Partial Observability}

\author{Jayawant~Bodagala and Balaji~Bodagala%
\thanks{Jayawant Bodagala and Balaji Bodagala are Independent researchers, CA, USA.}%
\thanks{Corresponding author: Jayawant Bodagala (e-mail: jayawantbodagala61@gmail.com).}%
}
\IEEEspecialpapernotice{
This work has been submitted to the IEEE for possible publication.
Copyright may be transferred without notice, after which this version may no longer be accessible.
}

\begin{document}
\maketitle


\begin{abstract}
Camera-based adaptive traffic signal control is inherently partially observable: detections can be missed, vehicle speeds and distances can be noisy, and a phase-change decision becomes temporally irreversible once yellow onset is initiated. This paper presents UCATSC, an interpretable uncertainty-aware constrained decision layer for vision-based adaptive signal control. UCATSC maintains a reduced movement-level belief state over queue, arrival, and service-age variables; evaluates admissible phase actions through finite-horizon counterfactual rollouts in belief space; and filters candidate actions using predictive dilemma-zone safety and service-age/starvation constraints before execution. The method is evaluated primarily in SUMO using matched seeds, classical baselines, a trained DQN-RL baseline, a safety-masked DQN variant, targeted safety/liveness/uncertainty stress tests, and a three-intersection corridor extension. In the tested scenarios, UCATSC demonstrates competitive mobility performance, eliminates observed dilemma-zone violations for the constrained variants, bounds service age in starvation stress tests, and maintains millisecond-level online runtime. A controlled physical vision testbed is included as supplementary feasibility evidence for the vision-to-belief interface; it is not presented as field validation of safety, emissions, or deployment readiness.
\end{abstract}


\section{Introduction}

Signalized intersections are critical bottlenecks in urban transportation networks. Their control policies affect delay, queue formation, stop--go behavior, fuel consumption, SUMO-modeled emission outputs, and safety-related surrogate measures \cite{Papageorgiou2003,Barth2009,RakhaAhn2004,Zhang2011}. Inefficient timing can increase idling and acceleration events near intersections, which can elevate local pollutant exposure for nearby pedestrians, drivers, and residents \cite{EPA2008,Karner2010NearRoad}. Adaptive signal control is therefore a practical target for improving mobility and reducing stop--go behavior, but deployment requires decisions that are explainable and robust to sensing uncertainty.

Classical adaptive traffic control systems such as SCOOT, SCATS, RHODES, OPAC, and max-pressure control demonstrate the value of feedback, optimization, and queue-pressure logic in signal timing \cite{Hunt1981SCOOT,Lowrie1982SCATS,MirchandaniHead2001,Gartner1983OPAC,Varaiya2013MaxPressure}. More recently, reinforcement learning (RL) methods have been proposed for traffic signal control because they can optimize long-horizon objectives and adapt to complex demand patterns \cite{DresnerStone2008,Wei2019PressLight,Belletti2022Nature}. However, both demand-responsive and learned controllers are commonly evaluated using deterministic state inputs such as queue length, vehicle count, occupancy, or image-derived features. This assumption is difficult to guarantee in camera-based deployments, where occlusion, illumination variation, motion blur, weather, detector confidence errors, and tracking failures make the true traffic state only partially observable \cite{SivaramanTrivedi2013,AzimjonovOzbayoglu2018,Kastrinaki2003Survey}.

This uncertainty matters because signal decisions are safety-critical and temporally irreversible. Once a controller initiates a phase transition, the yellow and all-red sequence must proceed according to timing rules; the controller cannot undo the action if a previously occluded vehicle appears in a dilemma-zone condition. Similarly, purely demand-driven policies can repeatedly prioritize dominant movements unless service and liveness constraints are explicitly checked. Reward shaping and post-hoc action masking can reduce these risks, but they do not by themselves provide an interpretable counterfactual check of both safety and starvation constraints before action execution \cite{Ng1999RewardShaping,Muller2022SafeMasking,Zhou2024ConstrainedRL}.

This paper addresses that deployment gap by formulating vision-based traffic signal control as a constrained decision problem under partial observability. UCATSC is not presented as a universally superior traffic signal controller. Instead, it is an interpretable uncertainty-aware constrained decision layer: it maintains a reduced movement-level belief state, evaluates candidate phase actions through finite-horizon belief-space rollouts, and executes only actions that satisfy predictive dilemma-zone and service-age constraints under the adopted reduced model. The central claim is therefore mechanism-oriented: explicit belief-space constraint checking changes the failure modes of adaptive signal control under uncertain camera-based observations.

\subsection{Contributions}
This paper proposes UCATSC as an uncertainty-aware constrained decision layer for adaptive signal control under vision-based partial observability. The contributions are:
\begin{itemize}
    \item \textbf{Reduced belief-space formulation for vision-based signal control:} We formulate the controller state as a movement-level belief over queue, arrival, and service-age variables rather than assuming perfectly observed queues or occupancies.
    \item \textbf{Counterfactual constrained phase-action checking:} We evaluate admissible phase actions through finite-horizon belief-space rollouts before execution, making the selected action explainable in terms of predicted queue burden, safety risk, and service age.
    \item \textbf{Predictive dilemma-zone safety filter:} We implement a belief-conditioned dilemma-zone risk constraint that blocks yellow onset when uncertain speed and distance estimates indicate unsafe stopping/clearing conditions.
    \item \textbf{Service-age liveness constraint and clean ablation evidence:} We impose service-age constraints to prevent indefinite under-service and include an S1X no-liveness ablation showing that removing liveness safeguards can produce severe minor-movement service-age growth.
    \item \textbf{Structured uncertainty and safety-masked learning tests:} We add a V5 bursty under-detection scenario and a DQN-RL safety-mask baseline to test whether uncertainty-aware belief correction and post-hoc masking reproduce the behavior of integrated constrained rollouts.
    \item \textbf{Reproducible SUMO evaluation with scoped claims:} We evaluate UCATSC using matched seeds, classical baselines, a representative trained DQN-RL baseline, targeted mechanism tests, SUMO-modeled per-vehicle emission outputs, runtime measurements, and a three-intersection corridor stress extension.
    \item \textbf{Supplementary physical vision-testbed feasibility evidence:} We retain a controlled camera-based testbed to demonstrate the vision-to-belief interface, while explicitly treating those results as feasibility evidence rather than field validation.
\end{itemize}

\subsection{Scope of Claims}
The paper separates two forms of evidence. SUMO experiments are used for standard performance evaluation, including delay, queue length, throughput, stops, fuel, emissions, and a three-intersection corridor spillback extension. The physical testbed is used to examine the feasibility of the vision-to-belief pipeline and the qualitative behavior of the controller under occlusion and degraded perception. Proxy results from the physical testbed are not interpreted as direct field emissions or crash-risk measurements.

\begin{table*}[!t]
\centering
\caption{Claim--evidence--limitation summary. The table scopes the paper's claims to the conducted SUMO and supplementary physical-testbed evidence.}
\label{tab:claim_evidence_limit}
\scriptsize
\setlength{\tabcolsep}{4pt}
\begin{adjustbox}{width=\textwidth}
\begin{tabularx}{\textwidth}{@{}p{0.23\textwidth}X X@{}}
\toprule
\textbf{Claim} & \textbf{Evidence in this paper} & \textbf{Limitation / scope} \\
\midrule
Predictive dilemma-zone safety filtering & UCATSC-det and UCATSC-full record zero observed dilemma-zone violations in the primary SUMO scenarios, while queue-greedy, max-pressure, DQN-RL, and UCATSC-no-safety record nonzero violation rates. V4 and V5 further test biased or bursty degraded observations. & The result is model-relative and simulation-based. It depends on the adopted dilemma-zone model, speed/distance uncertainty model, Monte Carlo estimator, and threshold $\epsilon$; it is not a field crash-risk guarantee. \\
Starvation/liveness protection & In S1, queue-greedy and max-pressure exceed the 90-s service-age stress threshold, while UCATSC-full remains below it. In S1X, UCATSC-no-liveness reaches 1880.85 s maximum E--W service age and 4101.35 s above the threshold, while UCATSC-full records zero starvation time. & The stress cases are designed SUMO scenarios. The guarantee is conditional on feasibility of the admissible action set and does not replace larger-network progression or offset coordination. \\
Structured vision-uncertainty handling & In V5, UCATSC-full reduces E--W delay and maximum E--W service age relative to UCATSC-det while maintaining zero violations under bursty E--W under-detection. & The improvement is movement-specific, not a universal aggregate-delay advantage. It depends on the assumed detection-probability and noise model. \\
Online runtime feasibility & UCATSC-full runs at millisecond-level mean and 95th-percentile computation time, well below the one-second control step in both isolated-intersection and corridor experiments. & Runtime is measured for the implemented reduced belief model and tested networks; larger networks or richer beliefs may require optimization. \\
Physical vision-testbed feasibility & The controlled physical testbed demonstrates zone-level perception, confidence-derived uncertainty, and the vision-to-belief interface under planned occlusion and lighting variation. & The testbed is supplementary feasibility evidence, not field deployment validation. Proxy risk and emission-linked measures are not direct field crash-risk or emissions measurements. \\
\bottomrule
\end{tabularx}
\end{adjustbox}
\end{table*}

\section{Related Work}

\subsection{Adaptive and Optimization-Based Signal Control}
Traditional traffic signal control includes fixed-time timing plans, actuated control, coordinated progression, and adaptive systems that adjust splits, offsets, and cycle lengths in response to measured demand \cite{Papageorgiou2003,Stevanovic2010Adaptive}. SCOOT and SCATS are representative deployed adaptive systems that use detector data to update timing decisions in real time \cite{Hunt1981SCOOT,Lowrie1982SCATS}. RHODES and OPAC introduced predictive and optimization-oriented signal control concepts that are conceptually related to model predictive control \cite{MirchandaniHead2001,Gartner1983OPAC}. Max-pressure control provides a theoretically motivated feedback policy based on queue pressure differences and has become an important non-RL benchmark for modern traffic control research \cite{Varaiya2013MaxPressure}.

\subsection{Reinforcement Learning for Traffic Signal Control}
RL-based traffic signal control has shown strong simulation performance, especially when accurate queue lengths, vehicle counts, or image-derived features are available. PressLight links deep RL with max-pressure concepts and is widely used as a strong learning-based benchmark \cite{Wei2019PressLight}. Large-scale traffic simulation platforms such as CityFlow have accelerated RL-based signal-control research by enabling reproducible training and evaluation across many intersections \cite{Zheng2019CityFlow}. Expert-level RL controllers have also demonstrated the potential of learned policies for complex signal environments \cite{Belletti2022Nature}. Under vision-based sensing, however, missed detections and occlusions can produce state aliasing: two different traffic situations may generate similar observed features. These limitations motivate explicit modeling of observation uncertainty and predictive safety constraints when phase transitions are irreversible.

\subsection{Partial Observability and Belief-Space Control}
Partially observable Markov decision processes provide a principled framework for decision-making when the true state is not directly observed \cite{Kaelbling1998POMDP}. In practice, exact POMDP solutions are intractable for realistic traffic systems because the microscopic state includes vehicle positions, speeds, intentions, and interactions. Belief-space control and probabilistic robotics methods address this challenge by maintaining an approximate belief over state variables that are relevant for control \cite{Thrun2005ProbRobotics}. UCATSC follows this direction by using a reduced movement-level belief state that preserves queue, arrival, and service-age uncertainty without attempting to solve the full microscopic POMDP.

\subsection{Safety, Interpretability, and Constrained Decision-Making}
Safety is often incorporated into learning systems using reward shaping, action masks, or constrained optimization \cite{Ng1999RewardShaping,Amodei2016AISafety,Muller2022SafeMasking,Zhou2024ConstrainedRL}. In traffic signal control, interpretability is important because agencies must justify timing decisions, diagnose failures, and certify safe operation under rare events \cite{Ault2020Interpretable,FHWA2008TSM}. Dilemma-zone behavior has a long history in traffic engineering, beginning with classical analysis of the amber signal problem and later work on yellow interval design \cite{Gazis1960Amber,LiuHermanGazis1996}. UCATSC combines these ideas by evaluating phase transitions through an explicit belief-conditioned risk constraint instead of relying only on a learned reward or a deterministic detector trigger.

\subsection{Positioning of UCATSC}
UCATSC is positioned between classical adaptive control, constrained MPC, and learning-based traffic signal control. Relative to classical adaptive and max-pressure controllers, UCATSC keeps explicit feedback and interpretable movement-level quantities but adds uncertainty-aware belief propagation and hard predictive safety/liveness checks. Relative to RL controllers, UCATSC does not learn a policy from reward alone; it evaluates candidate actions online using a reduced model and rejects actions that violate constraints. Relative to post-hoc action masking, UCATSC integrates safety and service-age constraints into the same counterfactual rollout used for mobility evaluation, rather than applying a one-step filter after a policy has already selected an action. Relative to full POMDP methods, UCATSC is deliberately approximate: it does not solve the microscopic POMDP, but uses a reduced movement-level belief that is computationally feasible at one-second resolution. Relative to generic constrained MPC, the novelty is the traffic-signal-specific integration of vision-derived uncertainty, dilemma-zone risk, service-age liveness, and admissible signal sequencing in a single counterfactual action-checking layer.

\subsection{Microsimulation for Reproducible Evaluation}
Microsimulation is commonly used to evaluate traffic signal controllers before field deployment because it provides controlled demand generation, repeatable random seeds, ground-truth traffic states, and standard performance metrics. In this paper, SUMO is selected as the main experimental platform because it supports microscopic traffic simulation, TraCI-based online control, route generation, queue and waiting-time outputs, and emission modeling \cite{Lopez2018SUMO}. The physical vision testbed is retained as a complementary sensing experiment rather than the sole basis for traffic performance claims.

\section{Problem Formulation}

\subsection{Partially Observable Signal-Control Model}
Let $x_t \in \cX$ denote the true but unobserved traffic state at time $t$, including vehicle positions, velocities, intentions, queue occupancies, and signal-relevant approach states. Let $u_t \in \cU$ denote a signal-control action such as extending the current green, terminating the current phase, or transitioning according to legal intergreen rules. Camera-based sensing produces a noisy observation
\begin{equation}
    z_t \in \cZ,\qquad z_t \sim p(z_t \mid x_t),
    \label{eq:obs_model}
\end{equation}
where $z_t$ may include detections, tracks, zone occupancies, motion states, and detection confidences.

The underlying problem can be viewed as a constrained partially observable stochastic decision process. The full Bayesian belief over microscopic state is
\begin{equation}
    b_t(x)=p(x_t=x\mid z_{1:t},u_{1:t-1}),
    \label{eq:full_belief}
\end{equation}
with the standard predict--update recursion
\begin{align}
    \bar{b}_{t+1}(x') &= \int p(x'\mid x,u_t)b_t(x)\,dx, \label{eq:belief_predict}\\
    b_{t+1}(x') &=
    \frac{p(z_{t+1}\mid x')\bar{b}_{t+1}(x')}
    {\int p(z_{t+1}\mid \tilde{x})\bar{b}_{t+1}(\tilde{x})\,d\tilde{x}}.
    \label{eq:belief_update}
\end{align}
Directly solving this POMDP is not tractable for online intersection control. UCATSC therefore uses a reduced belief over movement-level variables that are directly relevant to demand, safety, and service.

\subsection{Reduced Movement-Level Belief}
For each movement $m\in\cM$, UCATSC maintains
\begin{equation}
    b_t \triangleq \left\{p(Q_{m,t}),\;p(\lambda_{m,t}),\;\tau_{m,t}\right\}_{m\in\cM},
    \label{eq:belief_reduced}
\end{equation}
where $Q_{m,t}$ is queue length, $\lambda_{m,t}$ is arrival intensity, and $\tau_{m,t}$ is service age, defined as the time since movement $m$ was last served. Let
\begin{equation}
    y_t \triangleq (Q_{1,t},\ldots,Q_{M,t},\lambda_{1,t},\ldots,\lambda_{M,t},\tau_{1,t},\ldots,\tau_{M,t})
\end{equation}
denote the reduced aggregated state.

For each movement, queue evolution is modeled as
\begin{equation}
    Q_{m,t+1}=\max\left(0, Q_{m,t}+A_{m,t}-S_{m,t}(u_t)\right),
    \label{eq:queue_dyn}
\end{equation}
with arrivals
\begin{equation}
    A_{m,t}\sim \mathrm{Poisson}(\lambda_{m,t}\Delta t)
    \label{eq:arrival_model}
\end{equation}
and service bounded by
\begin{equation}
    0\le S_{m,t}(u_t)\le \mu_m\Delta t.
    \label{eq:service_bound}
\end{equation}
The service age evolves as
\begin{equation}
\tau_{m,t+1}=\begin{cases}
0, & \text{if movement }m\text{ is served under }u_t,\\
\tau_{m,t}+\Delta t, & \text{otherwise.}
\end{cases}
\label{eq:tau_update}
\end{equation}

\begin{figure}[!t]
    \centering
    \safeincludegraphics[width=\linewidth]{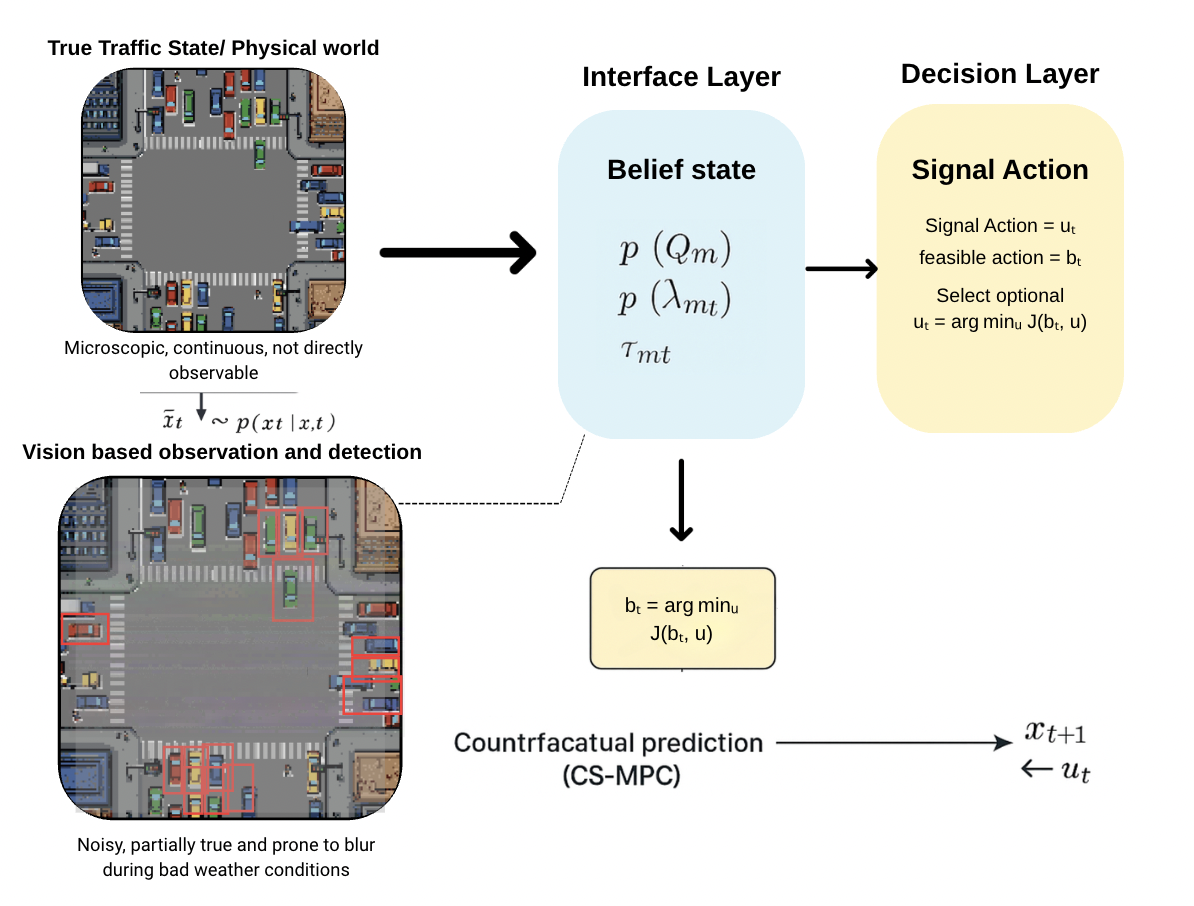}
    \caption{UCATSC decision architecture under partial observability. A noisy vision observation $z_t$ is converted into a reduced movement-level belief $b_t$. Candidate actions are rolled out in belief space and filtered by predictive safety and starvation constraints before execution.}
    \label{fig:pomdp_overview}
\end{figure}

\section{UCATSC Framework}

\subsection{Counterfactual Belief-Space Rollouts}
At each decision epoch, the controller constructs the legal action set $\cU_{\mathrm{adm}}(p_t)$ from the current phase $p_t$, minimum-green constraints, yellow/all-red rules, conflict constraints, and phase-sequencing rules. For each admissible candidate action $u$, UCATSC propagates the reduced belief over a horizon $H$:
\begin{equation}
    \hat{b}^{u}_{t+k}=\mathrm{Rollout}(b_t,u,k),\qquad k=0,\ldots,H.
    \label{eq:rollout_def}
\end{equation}
The predicted sequence is denoted
\begin{equation}
    \hat{b}^{u}_{t:t+H}=\{\hat{b}^{u}_{t},\hat{b}^{u}_{t+1},\ldots,\hat{b}^{u}_{t+H}\}.
    \label{eq:rollout_sequence}
\end{equation}
The mobility cost is
\begin{equation}
    J_Q(b_t,u)=\E\left[\sum_{k=0}^{H}\gamma^k D\left(\hat{b}^{u}_{t+k}\right)\right],
    \label{eq:cost}
\end{equation}
where $\gamma\in(0,1]$ and
\begin{equation}
    D(b)=\sum_{m\in\cM}\E_b[Q_m].
    \label{eq:delay_surrogate}
\end{equation}
Smoothness and phase-change penalties may be added as
\begin{equation}
    J(b_t,u)=J_Q(b_t,u)+\alpha J_{\Delta u}(u)+\beta J_{\mathrm{sw}}(u),
    \label{eq:aug_cost}
\end{equation}
but safety and starvation are treated as hard constraints rather than reward penalties.

\subsection{Predictive Dilemma-Zone Safety Constraint}
A yellow onset can be unsafe when an approaching vehicle cannot stop comfortably before the stop line and cannot clear the intersection before red. For a vehicle $i$, let $v_i$ be speed and $d_i$ be distance to the stop line. With reaction time $\delta_r$ and comfortable deceleration $a_{\max}$, the stopping distance is
\begin{equation}
    d_{\mathrm{stop}}(v_i)=v_i\delta_r+\frac{v_i^2}{2a_{\max}}.
    \label{eq:stop_distance}
\end{equation}
Stopping is feasible if $d_i\ge d_{\mathrm{stop}}(v_i)$. Let $L_i$ be the effective distance required to clear the intersection, $T_y$ the yellow interval, and $T_{ar}$ the all-red interval. Clearing is feasible if
\begin{equation}
    t_{\mathrm{clr}}(v_i,d_i)=\frac{L_i}{\max(v_i,v_{\min})}\le T_y+T_{ar}.
    \label{eq:clear_time}
\end{equation}
The dilemma-zone event for vehicle $i$ is
\begin{equation}
    \cD_i=\left(d_i<d_{\mathrm{stop}}(v_i)\right)\land
    \left(t_{\mathrm{clr}}(v_i,d_i)>T_y+T_{ar}\right).
    \label{eq:dilemma_event}
\end{equation}
For the set of relevant approaching vehicles $\cI_t$, the total event is
\begin{equation}
    \cD=\bigcup_{i\in\cI_t}\cD_i.
    \label{eq:dilemma_union}
\end{equation}
Because $(v_i,d_i)$ are uncertain under vision, the predictive risk is
\begin{equation}
    \mathcal{R}_{\mathrm{DZ}}\!\left(\hat{b}^{u}_{t+k}\right)=\Pp\left(\cD\mid \hat{b}^{u}_{t+k}\right).
    \label{eq:rdz}
\end{equation}
A candidate action that initiates yellow is admissible only if
\begin{equation}
    \mathcal{R}_{\mathrm{DZ}}\!\left(\hat{b}^{u}_{t+k}\right)\le \epsilon,
    \label{eq:rdz_constraint}
\end{equation}
where $\epsilon$ is the safety threshold.

\subsection{Starvation-Avoidance Constraint}
To prevent indefinite deprioritization of low-demand movements, UCATSC enforces
\begin{equation}
    \tau_{m,t+k}\le \tau_{\max},\qquad \forall m\in\cM,\; k=0,\ldots,H.
    \label{eq:starvation_constraint}
\end{equation}
This constraint acts as a liveness condition: even if one movement dominates the demand signal, the controller cannot allow another movement's service age to grow without bound.

\subsection{Belief-Space Constrained MPC Problem}
The resulting action-selection problem is
\begin{align}
    u_t^*\in \arg\min_{u\in\cU_{\mathrm{adm}}(p_t)}\quad & J(b_t,u) \label{eq:ucatsc_opt}\\
    \mathrm{s.t.}\quad 
    &\eqref{eq:queue_dyn},\eqref{eq:arrival_model},\eqref{eq:service_bound},\eqref{eq:tau_update} \text{ along rollout}, \nonumber\\
    &\mathcal{R}_{\mathrm{DZ}}\!\left(\hat{b}^{u}_{t+k}\right)\le \epsilon \text{ whenever yellow is initiated}, \nonumber\\
    &\tau_{m,t+k}\le \tau_{\max},\quad \forall m\in\cM,\; k=0,\ldots,H. \nonumber
\end{align}
The first feasible action with minimum rollout cost is executed, and the optimization is repeated at the next decision epoch.

\subsection{Conditional Constraint Properties}
The following statements formalize what the implemented constraint checks guarantee within the adopted reduced belief model. They are not field safety guarantees.

\noindent\textbf{Proposition 1 (model-relative predictive safety filtering):}
Suppose a yellow-onset action is executed only when the estimated dilemma-zone risk satisfies $\widehat{\mathcal{R}}_{\mathrm{DZ}}\le \epsilon$. Then, under the adopted reduced belief model, sensing likelihood, and Monte Carlo risk estimator, every executed yellow-onset action satisfies the model-relative predicted risk bound $\widehat{\mathcal{R}}_{\mathrm{DZ}}\le \epsilon$ at the decision epoch.

\noindent\emph{Justification:}
The controller evaluates the risk estimate for each candidate yellow-onset action and removes actions whose estimated risk exceeds $\epsilon$. Therefore, any yellow-onset action that remains eligible for execution satisfies the threshold by construction.

\noindent\textbf{Proposition 2 (rollout-horizon service-age feasibility):}
Suppose there exists at least one admissible action sequence within $\cU_{\mathrm{adm}}(p_t)$ whose predicted service ages satisfy $\tau_{m,t+k}\le\tau_{\max}$ for all movements $m$ and rollout steps $k=0,\ldots,H$. Then UCATSC selects an action from the feasible set that satisfies the service-age constraint over the rollout horizon.

\noindent\emph{Justification:}
UCATSC evaluates admissible actions by first rejecting candidates whose predicted rollouts violate the service-age constraint and then minimizing the rollout cost over the remaining feasible candidates. If the feasible set is nonempty, the selected action belongs to that set.

\noindent\textbf{Remark:}
Both properties are conditional on the reduced belief representation, sensing assumptions, risk estimator, threshold selection, signal timing rules, and feasibility of the constrained rollout problem. They bound the controller's predicted risk and predicted service age under the model; they do not certify real-world crash risk, detector correctness, or field deployment safety.

\begin{algorithm}[H]
\caption{UCATSC: Uncertainty-Aware Counterfactual Traffic Signal Control}
\label{alg:ucatsc}
\begin{algorithmic}[1]
\State \textbf{Inputs:} observation $z_t$, geometry $\cG$, phase set $\cP$, admissible-action rules, $H$, $\epsilon$, $\tau_{\max}$
\State \textbf{State:} reduced belief $b_t=\{p(Q_{m,t}),p(\lambda_{m,t}),\tau_{m,t}\}_{m\in\cM}$
\Loop
    \State Obtain observation $z_t$
    \State Update reduced belief $b_t$ using confidence and zone-level measurements
    \State Generate admissible actions $\cA\subseteq\cU_{\mathrm{adm}}(p_t)$
    \ForAll{$u\in\cA$}
        \State Roll out $\hat{b}^{u}_{t:t+H}$
        \State Evaluate dilemma-zone risk constraint \eqref{eq:rdz_constraint}
        \State Evaluate starvation constraint \eqref{eq:starvation_constraint}
        \State Compute rollout cost $J(b_t,u)$
    \EndFor
    \State Select lowest-cost feasible action $u_t^*$
    \State Execute $u_t^*$
\EndLoop
\end{algorithmic}
\end{algorithm}

\section{Implementation Details}

\subsection{Belief Representation}
In the SUMO experiments, the implementation uses a reduced aggregate belief over movements rather than a full microscopic particle filter. For each movement, queue length is represented as a finite discrete distribution over $\{0,1,\ldots,Q_{\max}\}$, arrival intensity is represented by an exponentially smoothed rate estimate with uncertainty inflation, and service age is represented deterministically. This representation is intentionally lightweight so that the controller can evaluate candidate actions online at one-second resolution.

The physical vision testbed uses the same movement-level interface, but its observations are derived from detections, tracks, zone occupancies, and detector confidence values. In the SUMO experiments, ground-truth states are first used to establish a reproducible benchmark. Vision uncertainty is then emulated through observation corruption models that randomly remove, delay, or perturb queue and approach-vehicle observations.

\subsection{Observation Uncertainty Model}
Let $\hat{Q}_{m,t}$ and $\hat{\lambda}_{m,t}$ denote observed queue and arrival-rate estimates. Under clean SUMO observation, these are obtained directly from simulator state. Under degraded sensing, observations are corrupted by a detection model:
\begin{equation}
    \tilde{n}_{m,t}\sim \mathrm{Binomial}(n_{m,t},p_{\mathrm{det}}),
    \label{eq:binomial_detection}
\end{equation}
where $n_{m,t}$ is the true number of observable vehicles in the relevant zone and $p_{\mathrm{det}}$ is the detection probability. Localization noise is injected into speed and distance estimates used by the dilemma-zone risk computation:
\begin{align}
    \tilde{v}_{i,t} &= v_{i,t}+\eta^v_{i,t}, \qquad \eta^v_{i,t}\sim \mathcal{N}(0,\sigma_v^2), \label{eq:speed_noise}\\
    \tilde{d}_{i,t} &= d_{i,t}+\eta^d_{i,t}, \qquad \eta^d_{i,t}\sim \mathcal{N}(0,\sigma_d^2). \label{eq:dist_noise}
\end{align}
The same likelihood model is used inside the belief update and the risk estimator, so degraded vision affects both demand estimation and yellow-onset feasibility.

For the structured V5 experiment, the detection probability is time varying and movement dependent:
\begin{equation}
p_{\mathrm{det},m}(t)=
\begin{cases}
p^{\mathrm{burst}}_{m}, & (t \bmod T_b)<T_{\mathrm{occ}},\\
p^{0}_{m}, & \text{otherwise},
\end{cases}
\label{eq:structured_detection}
\end{equation}
where $T_b$ is the burst period and $T_{\mathrm{occ}}$ is the occlusion-burst duration. Unlike independent random observation noise, this model creates repeated movement-specific under-detection. UCATSC-full applies a first-moment belief correction,
\begin{equation}
    \E[Q_{m,t}\mid \tilde{n}_{m,t},p_{\mathrm{det},m}]
    \approx
    \frac{\tilde{n}_{m,t}}{\max(p_{\mathrm{det},m},p_{\min})},
    \label{eq:belief_correction}
\end{equation}
with clipping by lane storage capacity. UCATSC-det does not apply this correction, allowing V5 to isolate the effect of movement-dependent perception bias.

\subsection{Monte Carlo Risk Estimation}
The implemented controller estimates $\mathcal{R}_{\mathrm{DZ}}$ by Monte Carlo sampling over uncertain speed and distance estimates. For each candidate yellow onset, $N_{\mathrm{MC}}$ samples of $(v_i,d_i)$ are drawn for relevant approaching vehicles, the event in \eqref{eq:dilemma_event} is evaluated for each sample, and risk is estimated as
\begin{equation}
    \widehat{\mathcal{R}}_{\mathrm{DZ}}=\frac{1}{N_{\mathrm{MC}}}
    \sum_{j=1}^{N_{\mathrm{MC}}}\mathbf{1}\!\left[
    \bigcup_{i\in\cI_t}\cD^{(j)}_i\right].
    \label{eq:mc_risk}
\end{equation}
A phase termination is infeasible when $\widehat{\mathcal{R}}_{\mathrm{DZ}}>\epsilon$.

\subsection{Controller Configuration}
Table~\ref{tab:implementation_params} lists the default parameters used by UCATSC in the SUMO experiments. These values should remain fixed across the main comparison unless a sensitivity study explicitly varies them.

\begin{table}[!t]
\centering
\caption{UCATSC implementation parameters used in the main SUMO experiments.}
\label{tab:implementation_params}
\footnotesize
\renewcommand{\arraystretch}{1.08}
\setlength{\tabcolsep}{4pt}
\begin{tabularx}{\columnwidth}{@{}>{\raggedright\arraybackslash}p{0.43\columnwidth}
                                >{\raggedright\arraybackslash}X@{}}
\toprule
\textbf{Parameter} & \textbf{Value / Description} \\
\midrule
Control step $\Delta t$        & 1 s \\
Rollout horizon $H$            & 30 steps \\
Discount factor $\gamma$       & 1.0 \\
Safety threshold $\epsilon$    & 0.05 \\
Maximum service age $\tau_{\max}$ & 120 s \\
Smoothness penalty $\alpha$    & 0.0 (not used in main experiments) \\
Phase-switch penalty $\beta$   & 4.0 \\
Minimum green $g_{\min}$       & 10 s \\
Maximum green $g_{\max}$       & 60 s \\
Yellow interval $T_y$          & 3 s \\
All-red interval $T_{ar}$      & 1 s \\
Queue belief support $Q_{\max}$ & Set by lane storage capacity \\
Risk computation               & Monte Carlo \\
Risk samples $N_{\mathrm{MC}}$ & 512 \\
Relevant vehicle lookahead     & 80 m \\
Episode duration $T_{\mathrm{ep}}$ & 3600 s \\
Random seeds per scenario      & 20 \\
\midrule
\multicolumn{2}{@{}p{\dimexpr\columnwidth-2\tabcolsep\relax}@{}}{\footnotesize Note: S1 and S1X use $\tau_{\max}=90$ s as a stress-test threshold.} \\
\bottomrule
\end{tabularx}
\end{table}

\subsection{Computational Complexity and Signal-Controller Compatibility}
At each decision epoch, UCATSC evaluates $|\cA|$ admissible actions over a horizon of $H$ steps. If the belief update uses $N_b$ support points per movement and there are $M$ movements, the approximate rollout complexity is
\begin{equation}
    O(|\cA|HN_bM).
    \label{eq:complexity}
\end{equation}
For an isolated four-leg intersection, $|\cA|$ is small because legal actions are restricted by the active phase, minimum-green rules, yellow clearance, all-red clearance, and conflict constraints. The framework can be mapped to NEMA-style or ring-barrier operation by defining $\cU_{\mathrm{adm}}(p_t)$ using permitted phase groups and barrier-crossing rules. In the corridor extension, the same admissible-action logic is applied locally at each intersection and augmented with downstream-link occupancy terms to discourage actions that push traffic into nearly saturated links.


\section{Experimental Setup}
\label{sec:experimental_setup}

\subsection{SUMO Microsimulation Benchmark}
The primary evaluation uses SUMO microsimulation with online signal control through TraCI \cite{Lopez2018SUMO}. The benchmark consists of an isolated four-leg signalized intersection with protected north--south and east--west phases. Each main simulation episode lasts 3600 s and uses a fixed random seed. Demand is generated from controlled route files so that all controllers are evaluated on matched traffic arrivals.

\begin{figure}[!t]
    \centering
    \safeincludegraphics[width=\linewidth]{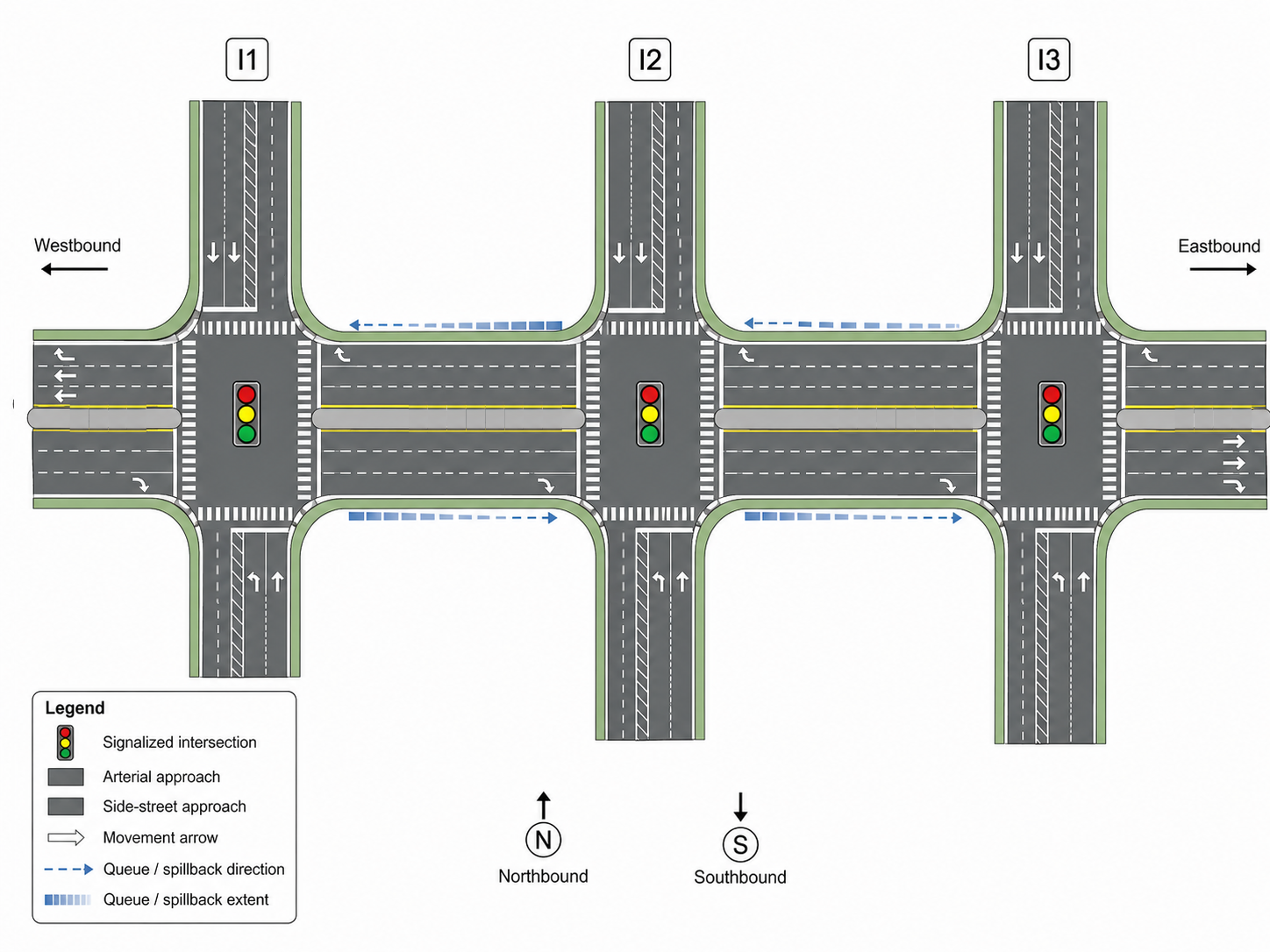}
    \caption{Three-intersection SUMO arterial-corridor benchmark used for the spillback and multi-signal evaluation. The schematic shows the three signalized intersections $I_1$--$I_3$, the main east--west arterial, north--south side-street approaches, lane-use arrows, crosswalks, and signalized junctions. Blue dashed arrows and shaded segments indicate the queue/spillback propagation direction and extent used to evaluate corridor-level control behavior.}
    \label{fig:sumo_network}
\end{figure}

\subsection{Demand, Sensing, and Stress-Test Scenarios}
The evaluation contains eight primary scenarios, two stress-test scenarios, and two targeted mechanism-isolation scenarios. Scenarios D1--D5 vary demand level and balance under clean observations; V1--V3 introduce mild, moderate, and severe sensing degradation. Scenario S1 is a starvation stress test with dominant north--south demand and persistent low east--west demand. Scenario V4 is an asymmetric degraded-vision stress test in which east--west approaches are under-detected relative to north--south approaches. The additional targeted scenarios S1X and V5 are used to isolate specific mechanisms: S1X removes ambiguity in the starvation ablation by testing a no-liveness variant, while V5 introduces bursty movement-dependent under-detection to test uncertainty-aware belief correction.

\begin{table}[!t]
\centering
\caption{SUMO scenario matrix for reproducible evaluation.}
\label{tab:scenario_matrix}
\footnotesize
\renewcommand{\arraystretch}{1.08}
\setlength{\tabcolsep}{3pt}
\begin{tabularx}{\columnwidth}{@{}>{\centering\arraybackslash}p{0.13\columnwidth}
                                >{\centering\arraybackslash}p{0.17\columnwidth}
                                >{\centering\arraybackslash}p{0.30\columnwidth}
                                >{\raggedright\arraybackslash}X@{}}
\toprule
\textbf{Scenario} &
\textbf{Demand} &
\textbf{Balance} &
\makecell[c]{\textbf{Observation}\\\textbf{quality}} \\
\midrule
D1 & Low    & Balanced              & Clean \\
D2 & Medium & Balanced              & Clean \\
D3 & High   & Balanced              & Clean \\
D4 & Medium & Unbalanced            & Clean \\
D5 & High   & Unbalanced            & Clean \\
V1 & Medium & Balanced              & Mild degradation \\
V2 & High   & Balanced              & Moderate degradation \\
V3 & High   & Balanced              & Severe degradation \\
S1 & High   & Dominant N--S demand  & Clean, service-age stress \\
V4 & Medium & Balanced              & Asymmetric E--W under-detection \\
\bottomrule
\end{tabularx}
\end{table}

In S1, the service-age threshold is set to $\tau_{\max}=90$ s to make liveness violations observable under dominant-flow demand. In V4, east--west observations are deliberately biased downward to test safety behavior under asymmetric vision degradation.

\begin{table}[!t]
\centering
\caption{Targeted mechanism-isolation scenarios added to the SUMO evaluation.}
\label{tab:targeted_scenarios}
\footnotesize
\setlength{\tabcolsep}{3pt}
\begin{tabularx}{\columnwidth}{@{}lYYX@{}}
\toprule
\textbf{Scenario} & \textbf{Purpose} & \textbf{Key stressor} & \textbf{Primary comparison} \\
\midrule
S1X & Liveness isolation & Dominant N--S flow with persistent low E--W demand & UCATSC-full vs. UCATSC-no-liveness \\
V5 & Uncertainty isolation & Bursty E--W under-detection with inflated speed/distance noise & UCATSC-full vs. UCATSC-det and safety-masked DQN \\
\bottomrule
\end{tabularx}
\end{table}

\subsection{Controllers Compared}
The SUMO experiments compare UCATSC with classical, optimization-based, learning-based, and ablation baselines:
\begin{itemize}
    \item \textbf{Fixed-time:} a static timing plan calibrated to nominal demand.
    \item \textbf{Queue-greedy:} a demand-responsive controller that prioritizes the movement group with the largest instantaneous queue estimate.
    \item \textbf{Max-pressure:} a feedback controller that selects phases based on queue-pressure differences \cite{Varaiya2013MaxPressure}.
    \item \textbf{DQN-RL:} a trained single-intersection deep Q-learning baseline with the same legal phase wrapper as the other online controllers.
    \item \textbf{UCATSC-det:} UCATSC with deterministic state estimates instead of uncertainty-aware belief.
    \item \textbf{UCATSC-no-safety:} UCATSC with the dilemma-zone constraint removed.
    \item \textbf{UCATSC-no-starvation:} UCATSC with the explicit service-age constraint removed while retaining legal phase sequencing and maximum-green rules.
    \item \textbf{UCATSC-full:} the full proposed method.
    \item \textbf{DQN-RL + safety mask:} the trained DQN-RL policy with the predictive dilemma-zone filter in \eqref{eq:dqn_mask}.
    \item \textbf{UCATSC-no-liveness:} a targeted ablation that removes explicit service-age forcing and maximum-green liveness protection while retaining minimum green, yellow, all-red, legal phase sequencing, and predictive dilemma-zone safety.
\end{itemize}

\subsection{DQN-RL Baseline Protocol}
To address the learning-based comparison gap, a trained DQN-RL baseline is included under the same SUMO network, demand scenarios, phase-transition wrapper, and held-out evaluation seeds. The DQN state contains normalized north--south and east--west queue estimates, the active phase, green age, and observation-quality indicators. The action set is binary: keep the current phase or request a legal transition to the competing phase, subject to minimum green, yellow, and all-red rules. The reward penalizes total queue burden, waiting behavior, and excessive switching. The DQN is trained offline for 500 episodes with 1800-s training episodes and an $\epsilon$-greedy exploration schedule decaying to 0.05. It is then evaluated without exploration on 20 matched seeds in each of the ten scenarios. The DQN baseline is intended as a representative trained RL comparison for an isolated intersection, not as a comprehensive benchmark against all modern RL traffic-signal methods. A safety-masked DQN variant is also used in the targeted experiments. Let $a^{\mathrm{DQN}}_t$ be the learned action and let $a^{\mathrm{keep}}$ denote keeping the current phase. The masked action is
\begin{equation}
a^{\mathrm{mask}}_t=
\begin{cases}
a^{\mathrm{keep}}, & a^{\mathrm{DQN}}_t\text{ requests yellow and }\widehat{\mathcal{R}}_{\mathrm{DZ}}>\epsilon,\\
a^{\mathrm{DQN}}_t, & \text{otherwise}.
\end{cases}
\label{eq:dqn_mask}
\end{equation}
This baseline tests whether a post-hoc predictive safety filter alone can match the integrated UCATSC rollout policy.

\begin{figure}[!t]
    \centering
    \safeincludegraphics[width=\linewidth]{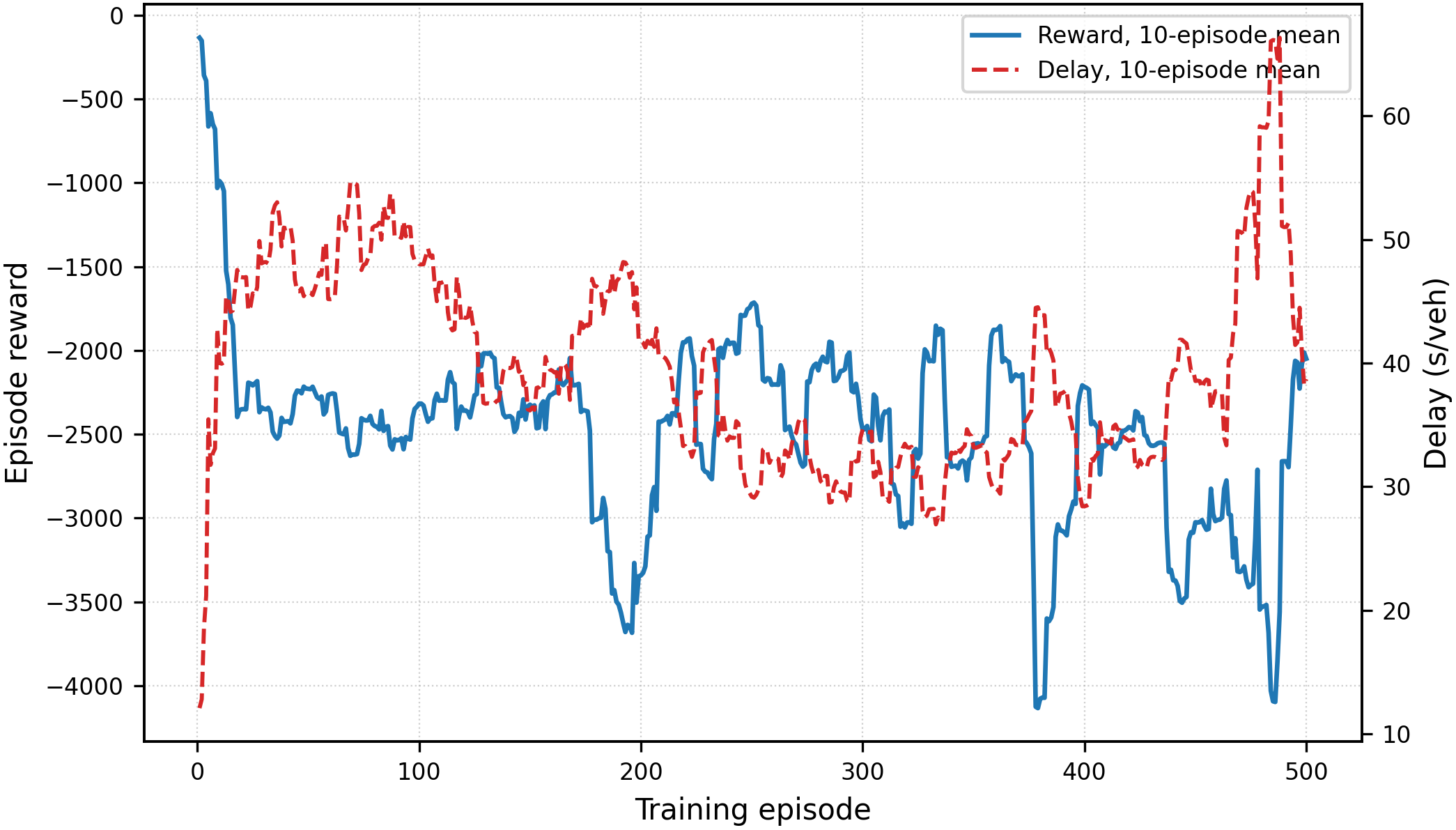}
    \caption{DQN-RL training trace over the 500-episode training protocol. The curve is included for transparency and shows the variability of a single learned policy across heterogeneous demand and degraded-sensing scenarios.}
    \label{fig:dqn_training}
\end{figure}

\subsection{Metrics and Statistical Reporting}
The primary SUMO metrics are average delay, average waiting time, average queue length, throughput, stops per vehicle, fuel consumption, CO$_2$/NO$_x$ emissions, dilemma-zone violation rate, service-age behavior, and runtime. The isolated-intersection dataset contains 1600 SUMO controller episodes: 1120 primary controller episodes from the original classical, UCATSC, and ablation controllers, 280 stress-test episodes for S1 and V4, and 200 held-out DQN-RL evaluation episodes across the ten isolated-intersection scenarios. The targeted S1X/V5 experiments add 320 isolated-intersection episodes, corresponding to two scenarios, eight controllers, and 20 matched seeds. The corridor extension adds 240 matched-seed controller episodes, corresponding to three corridor scenarios, four controllers, and 20 seeds. In total, the reported SUMO evidence covers 2160 controller episodes. Results are reported as means with 95\% bootstrap confidence intervals over matched runs unless otherwise stated.

\section{SUMO Results}
\label{sec:sumo_results}

\subsection{Result Ordering and Interpretation}
The SUMO results are organized around the mechanisms claimed by the paper rather than only by aggregate delay. The first results examine predictive dilemma-zone filtering, service-age/liveness behavior, structured vision uncertainty, and safety-masked learning. Aggregate mobility, SUMO-modeled per-vehicle emission outputs, scenario-wise delay, and runtime are then reported to show the mobility and computational cost associated with these constraints.

\subsection{Predictive Dilemma-Zone Safety Filtering}
Table~\ref{tab:safety_dz} summarizes dilemma-zone safety outcomes across the primary scenarios. Queue-greedy, max-pressure, DQN-RL, and the UCATSC-no-safety ablation produce nonzero violation rates. UCATSC-det, UCATSC-no-starvation, and UCATSC-full produce zero dilemma-zone violations in the completed primary SUMO runs. This result supports the central safety claim: explicit predictive safety filtering removes unsafe yellow-onset decisions observed in demand-driven or unconstrained variants.


\begin{table}[!t]
\centering
\caption{Dilemma-zone safety outcomes in the eight primary SUMO scenarios. Values are mean [95\% bootstrap CI].}
\label{tab:safety_dz}
\footnotesize
\renewcommand{\arraystretch}{1.05}
\setlength{\tabcolsep}{0pt}
\begin{tabular*}{0.78\columnwidth}{@{\extracolsep{\fill}} l r @{}}
\toprule
\textbf{Controller} & \textbf{DZ violations / 1000} \\
\midrule
Fixed-time           & 0.00 [0.00, 0.00] \\
Queue-greedy         & 1.44 [0.78, 2.20] \\
Max-pressure         & 1.61 [0.88, 2.49] \\
DQN-RL               & 0.53 [0.26, 0.82] \\
UCATSC-det           & 0.00 [0.00, 0.00] \\
UCATSC-no-safety     & 1.24 [0.64, 1.83] \\
UCATSC-no-starvation & 0.00 [0.00, 0.00] \\
UCATSC-full          & 0.00 [0.00, 0.00] \\
\bottomrule
\end{tabular*}
\end{table}

\subsection{Ablation Study}
Table~\ref{tab:ablation} reports the primary UCATSC ablation results. Removing the safety constraint slightly improves mean delay but introduces dilemma-zone violations. The deterministic and full UCATSC variants are close in mobility metrics, indicating that the primary demonstrated benefit in these experiments is the explicit constrained action filter rather than a large delay advantage from belief correction alone.

\begin{table}[!t]
\centering
\caption{UCATSC ablation summary over the primary scenarios. Values are mean [95\% bootstrap CI].}
\label{tab:ablation}
\scriptsize
\begin{tabular}{lcc}
\toprule
\textbf{Variant} & \textbf{Delay (s/veh)} & \textbf{DZ violations / 1000} \\
\midrule
UCATSC-det & 14.89 [14.34, 15.47] & 0.00 [0.00, 0.00] \\
UCATSC-no-safety & 14.86 [14.33, 15.44] & 1.24 [0.64, 1.83] \\
UCATSC-no-starvation & 14.87 [14.34, 15.46] & 0.00 [0.00, 0.00] \\
UCATSC-full & 14.88 [14.35, 15.48] & 0.00 [0.00, 0.00] \\
\bottomrule
\end{tabular}
\end{table}

\subsection{Starvation Stress Test}
\begin{figure}[!t]
    \centering
    \safeincludegraphics[width=\linewidth]{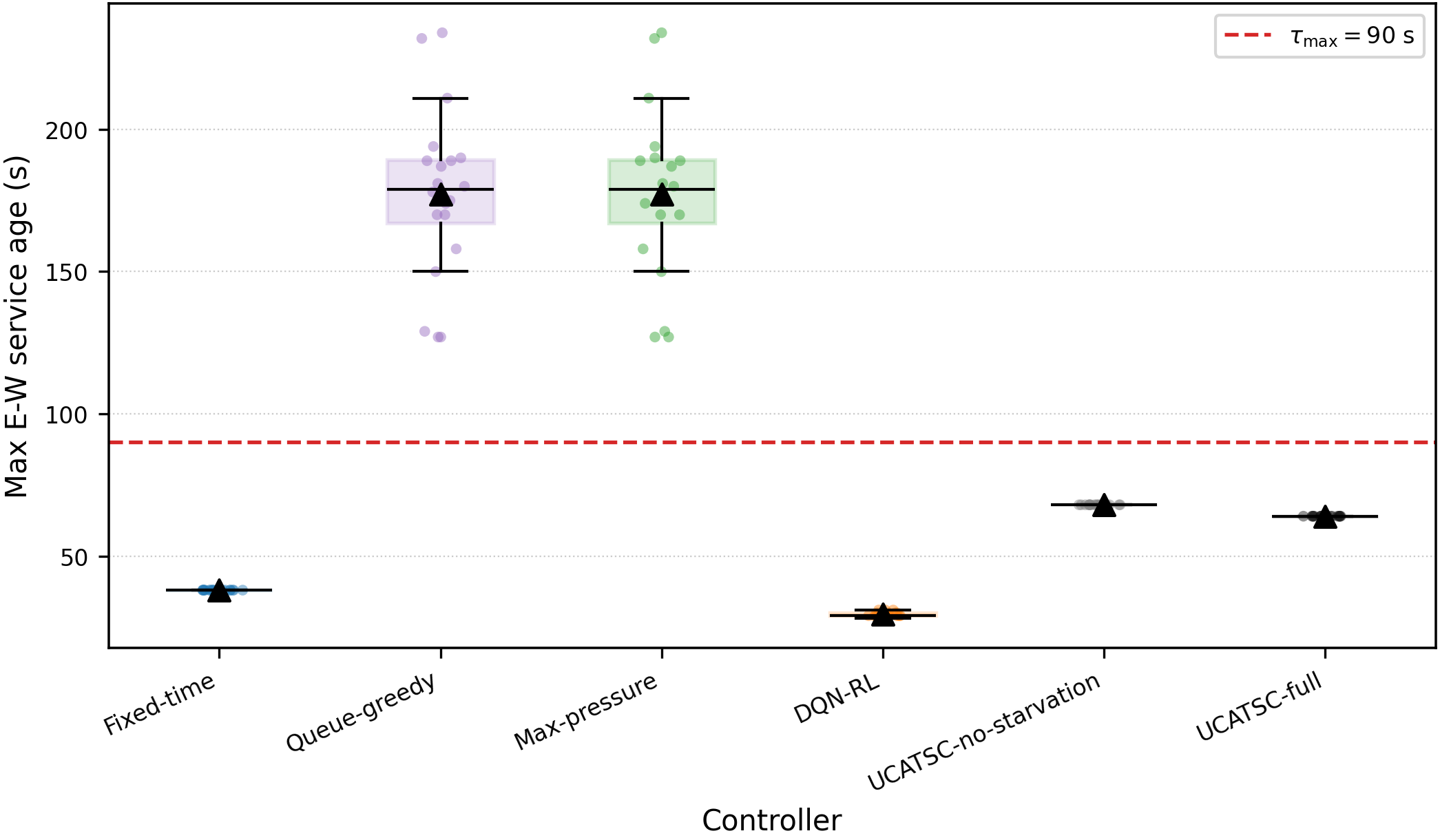}
    \caption{Distribution of maximum east--west service age in the starvation stress scenario, including DQN-RL. The dashed line marks the stress-test service-age threshold $\tau_{\max}=90$ s.}
    \label{fig:starvation_service_age}
\end{figure}

Table~\ref{tab:starvation_stress} and Fig.~\ref{fig:starvation_service_age} report the S1 starvation stress scenario. Queue-greedy and max-pressure reach a mean maximum east--west service age of 177.25 s and spend 175.40 s above the 90 s stress-test service-age limit, with approximately 5.05 starvation episodes per run. UCATSC-full bounds the maximum east--west service age at 64.00 s and records zero time above the service-age limit. DQN-RL avoids starvation in this stress setting but does so with substantially higher delay than UCATSC-full. The no-starvation ablation also remains below the threshold because the implementation retains legal phase sequencing and maximum-green timing. Therefore, S1 validates service protection in the UCATSC timing structure, while the S1X experiment below isolates the additional role of explicit liveness safeguards more cleanly.

\begin{table*}[!t]
\centering
\caption{Starvation stress scenario under dominant north--south demand. Values are mean [95\% bootstrap CI]. The service-age threshold is $\tau_{\max}=90$ s for this stress test.}
\label{tab:starvation_stress}
\scriptsize
\renewcommand{\arraystretch}{1.08}
\setlength{\tabcolsep}{3pt}
\begin{tabular*}{\textwidth}{@{\extracolsep{\fill}}lcccc@{}}
\toprule
\textbf{Controller} & \textbf{Delay (s/veh)} & \textbf{Max E-W age (s)} & \textbf{Time above limit (s)} & \textbf{Starvation events} \\
\midrule
Fixed-time & 41.87 [38.65, 45.27] & 38.00 [38.00, 38.00] & 0.00 [0.00, 0.00] & 0.00 [0.00, 0.00] \\
Queue-greedy & 14.74 [14.17, 15.30] & 177.25 [164.35, 190.16] & 175.40 [150.30, 200.45] & 5.05 [4.25, 5.80] \\
Max-pressure & 14.74 [14.17, 15.30] & 177.25 [164.35, 190.16] & 175.40 [150.30, 200.45] & 5.05 [4.25, 5.80] \\
DQN-RL & 36.90 [36.35, 37.51] & 29.45 [29.15, 29.85] & 0.00 [0.00, 0.00] & 0.00 [0.00, 0.00] \\
UCATSC-no-starvation & 11.44 [11.31, 11.56] & 68.00 [68.00, 68.00] & 0.00 [0.00, 0.00] & 0.00 [0.00, 0.00] \\
UCATSC-full & 11.30 [11.21, 11.38] & 64.00 [64.00, 64.00] & 0.00 [0.00, 0.00] & 0.00 [0.00, 0.00] \\
\bottomrule
\end{tabular*}
\end{table*}

\subsection{Asymmetric Degraded-Vision Safety Stress Test}
\begin{figure}[!t]
    \centering
    \safeincludegraphics[width=\linewidth]{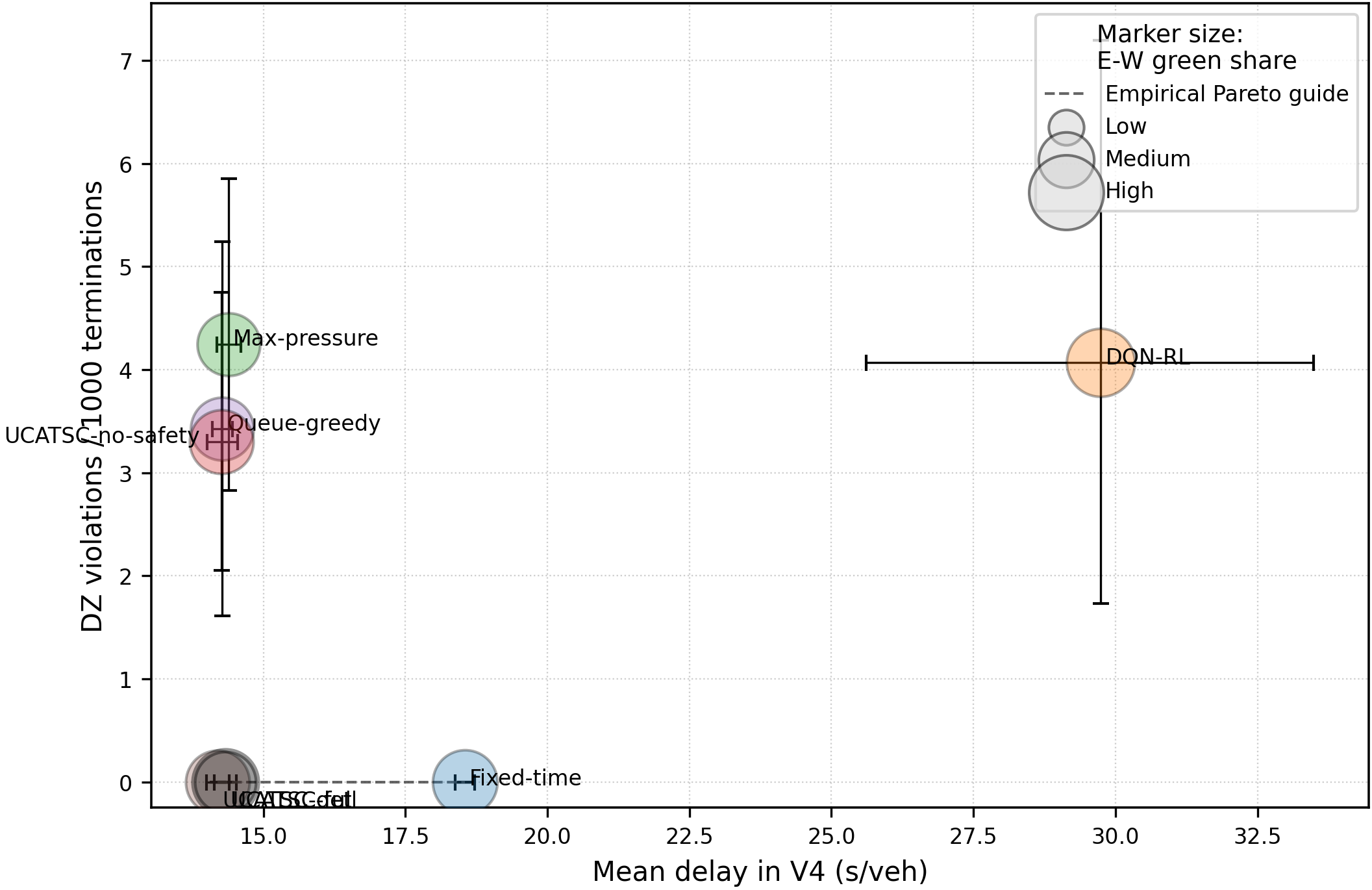}
    \caption{Delay--safety tradeoff in the asymmetric degraded-vision scenario. Marker size indicates the east--west green share.}
    \label{fig:v4_delay_safety}
\end{figure}

Table~\ref{tab:asymmetric_vision} and Fig.~\ref{fig:v4_delay_safety} report the V4 asymmetric occlusion scenario. Queue-greedy and max-pressure remain competitive in delay but produce nonzero dilemma-zone violation rates. DQN-RL performs poorly in this stress case, with a mean delay of 29.74 s/veh and 4.07 violations per 1000 terminations. UCATSC-full produces zero violations with only a small mobility penalty relative to UCATSC-det. These results show that the safety filter is active under biased observations and that a trained unconstrained RL baseline does not by itself provide the same safety guarantee.

\begin{table*}[!t]
\centering
\caption{Asymmetric degraded-vision stress scenario. East--west approaches are under-detected relative to north--south approaches. Values are mean [95\% bootstrap CI].}
\label{tab:asymmetric_vision}
\scriptsize
\renewcommand{\arraystretch}{1.08}
\setlength{\tabcolsep}{3pt}
\begin{tabular*}{\textwidth}{@{\extracolsep{\fill}}lcccc@{}}
\toprule
\textbf{Controller} & \textbf{Delay (s/veh)} & \textbf{E-W delay (s/veh)} & \textbf{E-W green share} & \textbf{DZ viol./1000} \\
\midrule
Queue-greedy & 14.28 [14.08, 14.47] & 15.17 [14.95, 15.40] & 0.466 [0.464, 0.468] & 3.42 [1.81, 5.24] \\
Max-pressure & 14.38 [14.17, 14.58] & 15.28 [14.97, 15.58] & 0.464 [0.462, 0.466] & 4.25 [2.64, 5.85] \\
DQN-RL & 29.74 [25.82, 33.77] & 18.08 [16.02, 20.10] & 0.569 [0.534, 0.605] & 4.07 [1.69, 6.85] \\
UCATSC-det & 14.20 [14.00, 14.40] & 14.82 [14.52, 15.12] & 0.483 [0.479, 0.486] & 0.00 [0.00, 0.00] \\
UCATSC-no-safety & 14.26 [14.00, 14.55] & 14.90 [14.49, 15.33] & 0.483 [0.479, 0.487] & 3.30 [2.04, 4.75] \\
UCATSC-full & 14.33 [14.12, 14.53] & 14.86 [14.58, 15.17] & 0.479 [0.477, 0.482] & 0.00 [0.00, 0.00] \\
\bottomrule
\end{tabular*}
\end{table*}

\subsection{Targeted Mechanism-Isolation Experiments}
\label{sec:targeted_mechanism}

The targeted S1X and V5 experiments isolate two reviewer-critical mechanisms more directly: whether the liveness constraint matters beyond legal timing, and whether uncertainty-aware belief helps when observation errors are structured rather than independent. The S1X and V5 experiments address these mechanisms.

\subsubsection{Clean liveness ablation}
\begin{figure}[!t]
    \centering
    \safeincludegraphics[width=\linewidth]{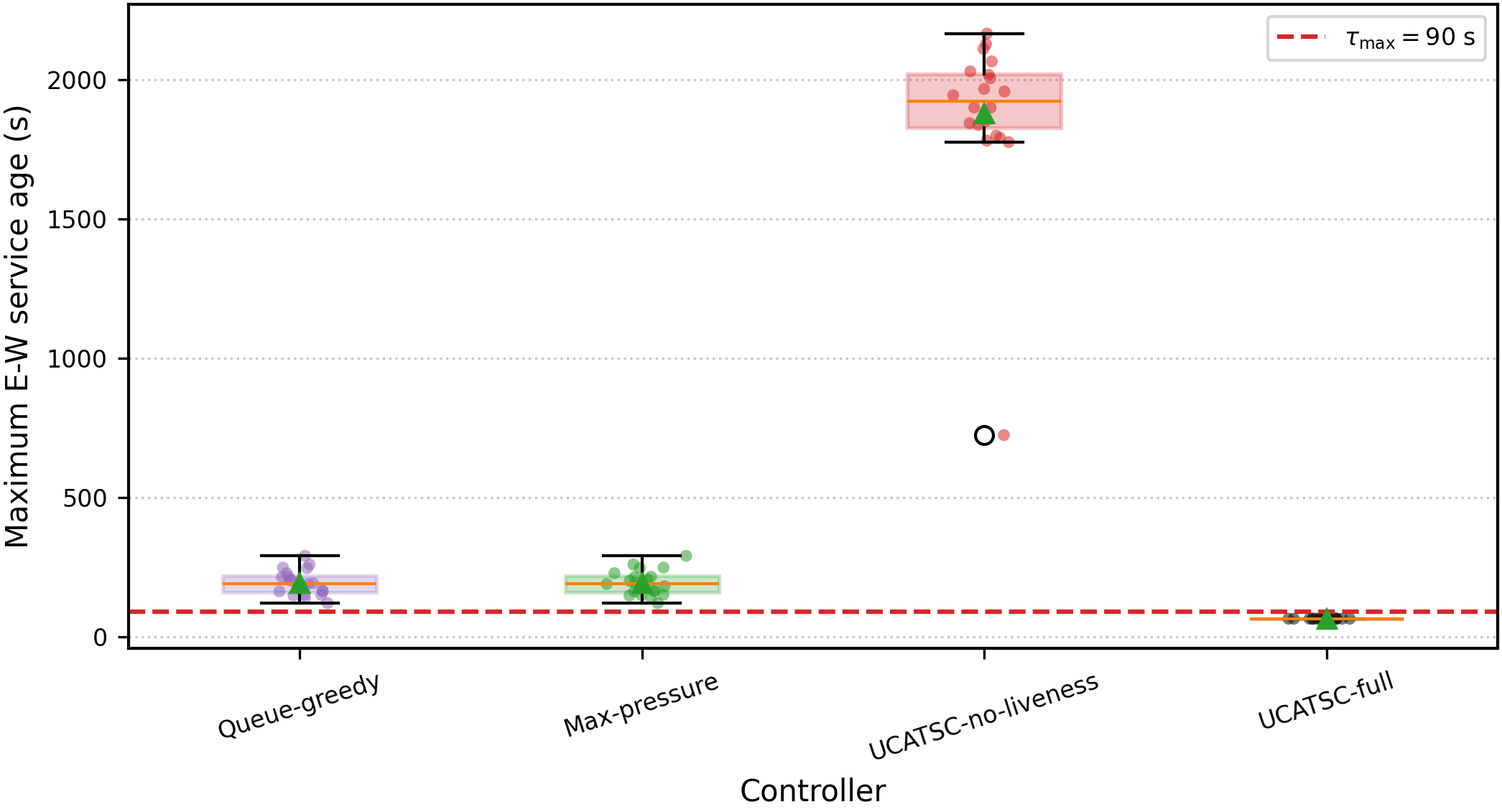}
    \caption{Clean S1X liveness ablation. UCATSC-no-liveness removes explicit service-age forcing and maximum-green liveness protection while retaining legal signal sequencing and dilemma-zone safety.}
    \label{fig:targeted_liveness}
\end{figure}

Table~\ref{tab:targeted_liveness} and Fig.~\ref{fig:targeted_liveness} show the S1X liveness-ablation result. UCATSC-no-liveness removes explicit service-age forcing and maximum-green liveness protection while retaining minimum green, yellow, all-red, legal sequencing, and dilemma-zone safety. This variant reaches a mean maximum E--W service age of 1880.85 s and spends 4101.35 s above the 90-s liveness threshold. UCATSC-full keeps the maximum E--W service age at 64.00 s with zero starvation time and zero starvation events. This is the cleanest evidence that the liveness mechanism is not merely cosmetic: when liveness safeguards are removed, the minor movement can be systematically under-served even though the phase sequence remains legally valid.

\begin{table*}[!t]
\centering
\caption{Targeted S1X liveness-ablation results. Values are mean [95\% bootstrap CI] over 20 matched seeds.}
\label{tab:targeted_liveness}
\scriptsize
\setlength{\tabcolsep}{4pt}
\begin{adjustbox}{width=\textwidth}
\begin{tabular}{lccccc}
\toprule
\textbf{Controller} & \textbf{Delay (s/veh)} & \textbf{E--W delay (s/veh)} & \textbf{Max E--W age (s)} & \textbf{Starv. time (s)} & \textbf{Starv. events} \\
\midrule
Queue-greedy & 15.71 [15.10, 16.33] & 12.31 [11.98, 12.65] & 193.00 [172.39, 212.05] & 289.75 [233.29, 349.72] & 7.30 [6.45, 8.20] \\
Max-pressure & 15.71 [15.10, 16.33] & 12.31 [11.98, 12.65] & 193.00 [172.39, 212.05] & 289.75 [233.29, 349.72] & 7.30 [6.45, 8.20] \\
UCATSC-no-liveness & 15.51 [15.25, 15.77] & 194.77 [186.79, 203.13] & 1880.85 [1736.14, 1978.87] & 4101.35 [3938.75, 4205.00] & 12.15 [11.65, 12.65] \\
UCATSC-full & 12.13 [11.98, 12.30] & 32.26 [31.52, 33.11] & 64.00 [64.00, 64.00] & 0.00 [0.00, 0.00] & 0.00 [0.00, 0.00] \\
\bottomrule
\end{tabular}
\end{adjustbox}
\end{table*}

\subsubsection{Structured uncertainty and safety-masked learning}
\begin{figure}[!t]
    \centering
    \safeincludegraphics[width=\linewidth]{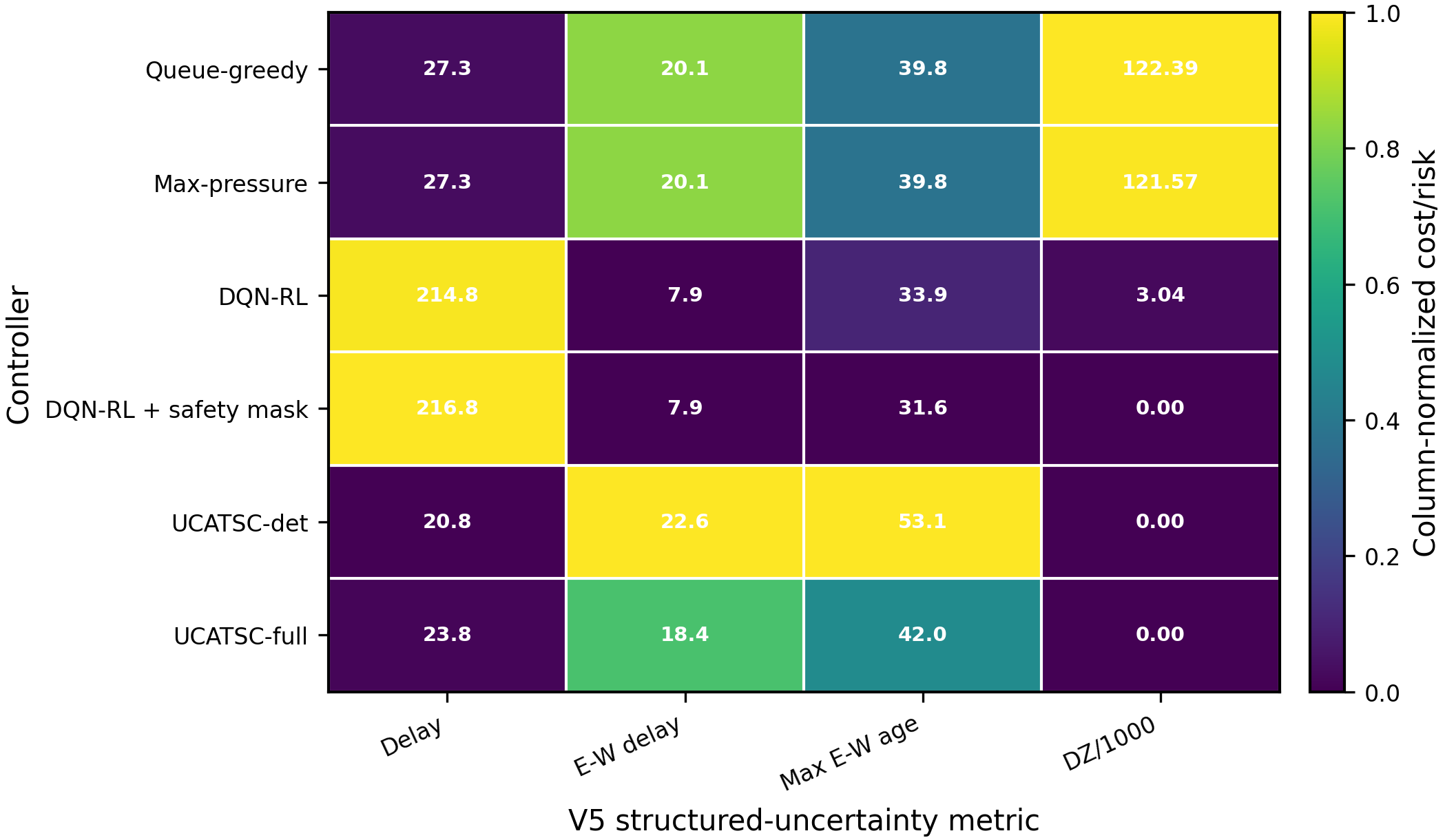}
    \caption{V5 structured-uncertainty metric matrix. Columns are normalized independently for visualization; cell text reports the original metric values.}
    \label{fig:v5_uncertainty_matrix}
\end{figure}

\begin{figure}[!t]
    \centering
    \safeincludegraphics[width=\linewidth]{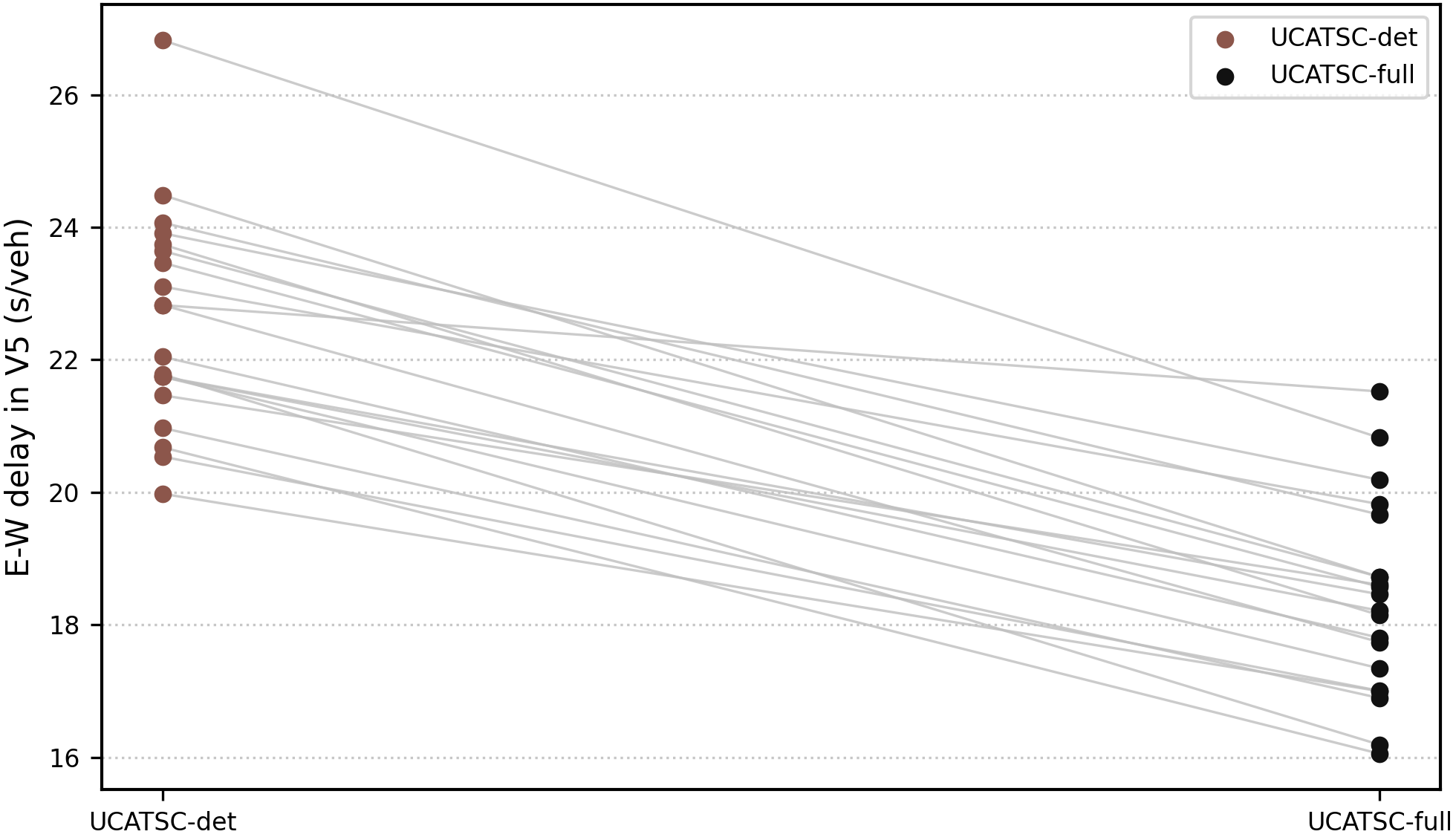}
    \caption{Paired-seed comparison of UCATSC-det and UCATSC-full for E--W delay in V5. The scenario targets the under-detected movement rather than only aggregate delay.}
    \label{fig:v5_full_det_pair}
\end{figure}

Table~\ref{tab:targeted_uncertainty} and Figs.~\ref{fig:v5_uncertainty_matrix}--\ref{fig:v5_full_det_pair} report V5. Queue-greedy and max-pressure produce more than 120 dilemma-zone violations per 1000 terminations, showing that demand-responsive phase changes can be unsafe under structured sensing degradation. DQN-RL has nonzero violations and very high aggregate delay. UCATSC-det and UCATSC-full both produce zero violations, but their mobility tradeoff differs. UCATSC-det has lower aggregate delay, whereas UCATSC-full reduces E--W delay from 22.58 s/veh to 18.37 s/veh and reduces maximum E--W service age from 53.15 s to 42.05 s. Thus, V5 supports a specific uncertainty claim: belief correction improves the under-detected movement under structured asymmetric occlusion while preserving safety, rather than guaranteeing lower aggregate delay in every degraded-vision case.

\begin{table*}[!t]
\centering
\caption{Targeted V5 structured bursty asymmetric-occlusion results. Values are mean [95\% bootstrap CI] over 20 matched seeds.}
\label{tab:targeted_uncertainty}
\scriptsize
\setlength{\tabcolsep}{4pt}
\begin{adjustbox}{width=\textwidth}
\begin{tabular}{lccccc}
\toprule
\textbf{Controller} & \textbf{Delay (s/veh)} & \textbf{E--W delay (s/veh)} & \textbf{E--W green share} & \textbf{DZ viol./1000} & \textbf{Max E--W age (s)} \\
\midrule
Queue-greedy & 27.27 [26.11, 28.57] & 20.11 [19.11, 21.17] & 0.475 [0.472, 0.477] & 122.39 [109.45, 134.57] & 39.80 [37.85, 41.60] \\
Max-pressure & 27.27 [26.11, 28.57] & 20.11 [19.11, 21.17] & 0.475 [0.472, 0.477] & 121.57 [109.62, 133.67] & 39.80 [37.85, 41.60] \\
DQN-RL & 214.83 [210.13, 219.58] & 7.90 [7.65, 8.14] & 0.829 [0.825, 0.833] & 3.04 [1.33, 5.10] & 33.90 [31.70, 36.05] \\
DQN-RL + safety mask & 216.83 [211.24, 222.45] & 7.93 [7.66, 8.22] & 0.830 [0.827, 0.834] & 0.00 [0.00, 0.00] & 31.65 [29.05, 34.75] \\
UCATSC-det & 20.84 [20.19, 21.53] & 22.58 [21.93, 23.31] & 0.423 [0.419, 0.428] & 0.00 [0.00, 0.00] & 53.15 [50.35, 55.95] \\
UCATSC-full & 23.81 [22.93, 24.79] & 18.37 [17.77, 19.03] & 0.467 [0.464, 0.469] & 0.00 [0.00, 0.00] & 42.05 [39.95, 44.00] \\
\bottomrule
\end{tabular}
\end{adjustbox}
\end{table*}

\begin{figure}[!t]
    \centering
    \safeincludegraphics[width=\linewidth]{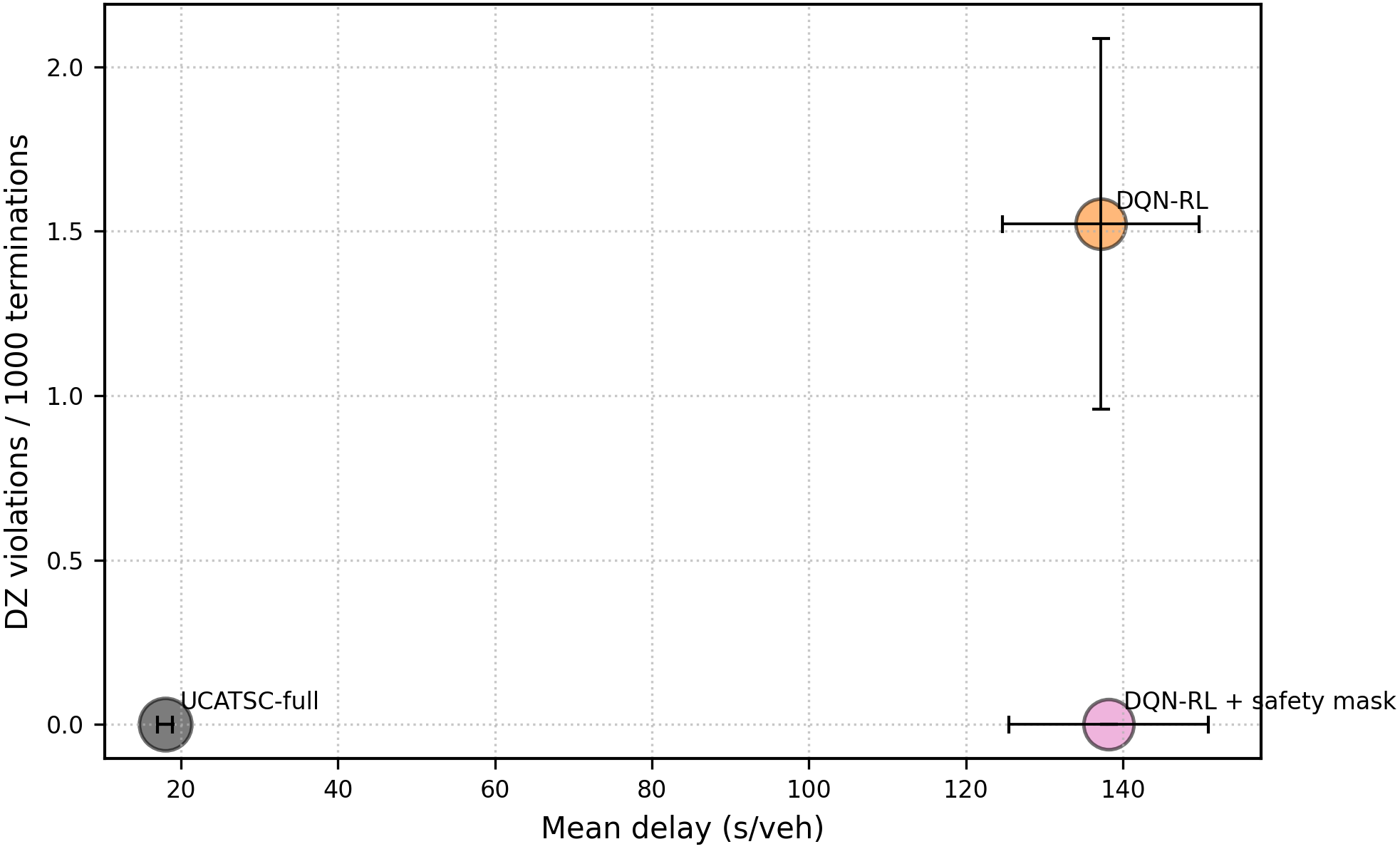}
    \caption{Targeted delay--safety tradeoff for DQN-RL, safety-masked DQN-RL, and UCATSC-full. Marker size is proportional to throughput.}
    \label{fig:dqn_mask_tradeoff}
\end{figure}

Table~\ref{tab:dqn_mask} and Fig.~\ref{fig:dqn_mask_tradeoff} compare DQN-RL, DQN-RL with the predictive safety mask, and UCATSC-full over the targeted scenarios. The mask eliminates DQN dilemma-zone violations, reducing the mean rate from 1.52 to 0.00 violations per 1000 terminations. However, it does not make the learned policy competitive: masked DQN has 138.18 s/veh mean delay and lower throughput, while UCATSC-full has 17.97 s/veh mean delay, higher throughput, and zero violations. The result distinguishes a post-hoc safety filter from an integrated constrained rollout policy.

\begin{table}[!t]
\centering
\caption{Safety-masked DQN comparison over targeted S1X/V5 scenarios. Values are mean [95\% bootstrap CI].}
\label{tab:dqn_mask}
\scriptsize
\setlength{\tabcolsep}{2.3pt}
\begin{tabular}{lccc}
\toprule
\textbf{Controller} & \textbf{Delay} & \textbf{Throughput} & \textbf{DZ/1000} \\
\midrule
DQN-RL & 137.19 [113.35, 161.33] & 1712.6 [1660.5, 1764.7] & 1.52 [0.51, 2.66] \\
DQN-RL + mask & 138.18 [113.75, 162.60] & 1709.3 [1654.6, 1762.3] & 0.00 [0.00, 0.00] \\
UCATSC-full & 17.97 [16.15, 19.80] & 1934.2 [1911.4, 1956.6] & 0.00 [0.00, 0.00] \\
\bottomrule
\end{tabular}
\end{table}

\subsection{Aggregate Mobility Performance}
Table~\ref{tab:sumo_main_metrics} reports the main SUMO traffic-performance metrics aggregated across the eight primary demand and sensing scenarios. UCATSC-full achieves the lowest delay among the main evaluated controllers, with 14.88 s/veh compared with 19.16 s/veh for fixed-time, 16.15 s/veh for queue-greedy, 16.18 s/veh for max-pressure, and 18.56 s/veh for DQN-RL. UCATSC-full also preserves the matched-seed throughput achieved by the classical baselines. The DQN-RL baseline improves mean delay relative to fixed-time but serves fewer vehicles in several scenarios and produces nonzero dilemma-zone violations, so its delay should not be interpreted without the accompanying throughput and safety columns.

\begin{table*}[!t]
\centering
\caption{Main SUMO traffic-performance results including a trained DQN-RL baseline. Values are mean [95\% bootstrap CI] over the eight primary scenarios and matched seeds.}
\label{tab:sumo_main_metrics}
\scriptsize
\renewcommand{\arraystretch}{1.08}
\setlength{\tabcolsep}{3pt}
\begin{tabular*}{\textwidth}{@{\extracolsep{\fill}}lccccc@{}}
\toprule
\textbf{Controller} & \textbf{Delay (s/veh)} & \textbf{Waiting (s/veh)} & \textbf{Queue (veh)} & \textbf{Throughput (veh)} & \textbf{DZ viol./1000} \\
\midrule
Fixed-time & 19.16 [18.83, 19.48] & 10.73 [10.57, 10.90] & 4.22 [3.98, 4.44] & 1415.8 [1350.6, 1480.9] & 0.00 [0.00, 0.00] \\
Queue-greedy & 16.15 [15.49, 16.81] & 6.67 [6.34, 7.00] & 2.78 [2.55, 3.02] & 1415.8 [1350.6, 1480.9] & 1.44 [0.78, 2.20] \\
Max-pressure & 16.18 [15.51, 16.85] & 6.69 [6.36, 7.03] & 2.79 [2.56, 3.02] & 1415.8 [1350.6, 1480.9] & 1.61 [0.88, 2.49] \\
DQN-RL & 18.56 [17.98, 19.13] & 8.85 [8.57, 9.14] & 2.64 [2.43, 2.84] & 1061.1 [987.2, 1133.9] & 0.53 [0.26, 0.82] \\
UCATSC-det & 14.89 [14.34, 15.47] & 6.13 [5.85, 6.42] & 2.54 [2.34, 2.75] & 1415.8 [1350.6, 1480.9] & 0.00 [0.00, 0.00] \\
UCATSC-full & 14.88 [14.35, 15.48] & 6.12 [5.85, 6.41] & 2.54 [2.34, 2.75] & 1415.8 [1350.6, 1480.9] & 0.00 [0.00, 0.00] \\
\bottomrule
\end{tabular*}
\end{table*}

\begin{figure}[!t]
    \centering
    \safeincludegraphics[width=\linewidth]{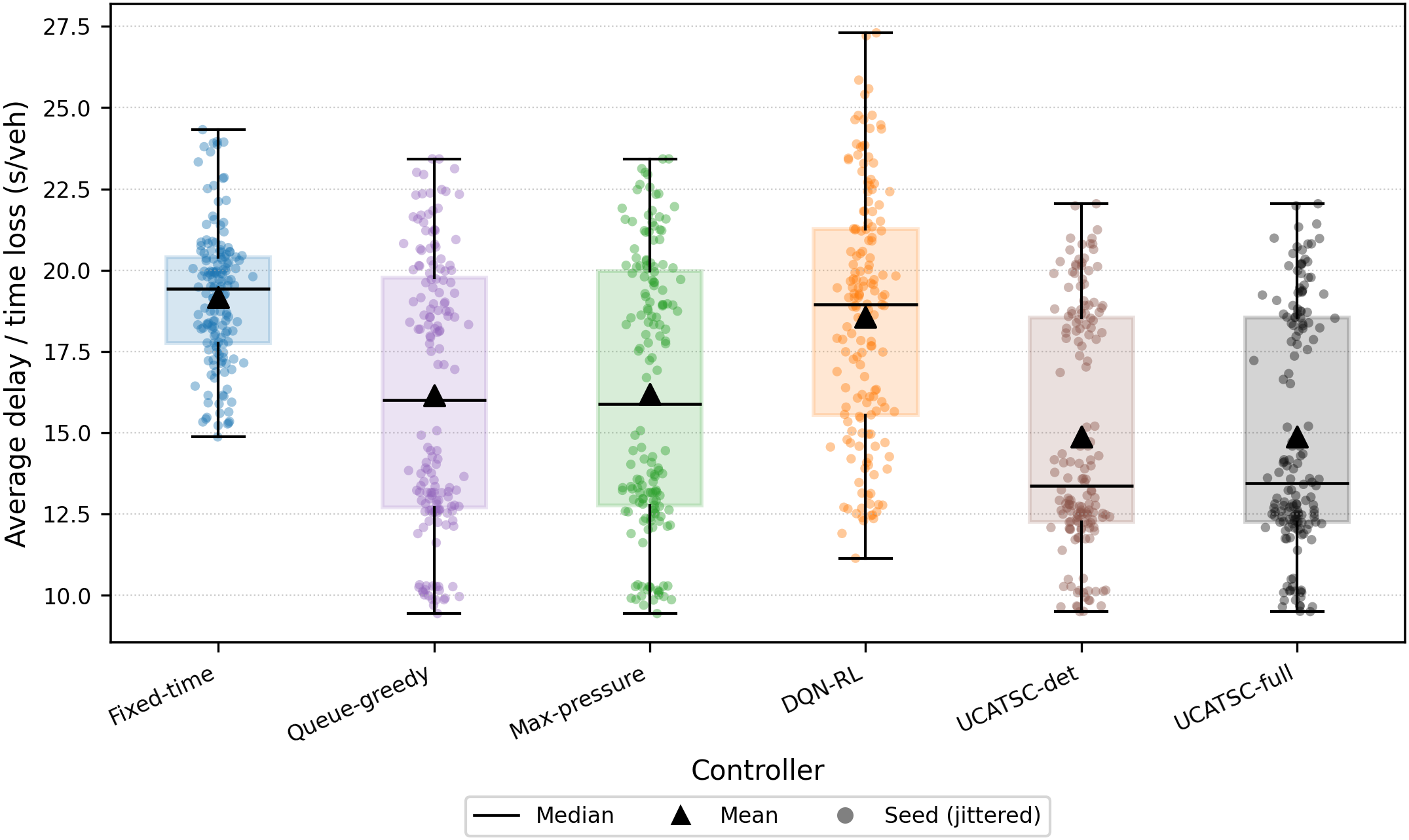}
    \caption{SUMO average-delay distribution including the trained DQN-RL baseline. Each distribution summarizes matched held-out seeds across the eight primary scenarios.}
    \label{fig:sumo_delay_box}
\end{figure}

\subsection{SUMO-Modeled Emission and Stop--Go Outputs}
Table~\ref{tab:sumo_emissions} reports per-arrived-vehicle SUMO-modeled emission and stop--go outputs. Per-vehicle normalization is used because DQN-RL does not preserve the same throughput as the other controllers. UCATSC-full has the lowest mean SUMO-modeled fuel, CO$_2$, and NO$_x$ outputs per arrived vehicle among the compared controllers. DQN-RL remains close to fixed-time in SUMO-modeled per-vehicle emission outputs and produces more stops per vehicle than UCATSC-full, consistent with its less stable phase-selection behavior under the fixed training protocol.

\begin{table*}[!t]
\centering
\caption{Per-arrived-vehicle SUMO-modeled emission and stop--go results in SUMO. Values are mean [95\% bootstrap CI]. Per-vehicle normalization is used because the DQN-RL baseline serves fewer vehicles in several scenarios.}
\label{tab:sumo_emissions}
\scriptsize
\renewcommand{\arraystretch}{1.08}
\setlength{\tabcolsep}{3pt}
\begin{tabular*}{\textwidth}{@{\extracolsep{\fill}}lcccc@{}}
\toprule
\textbf{Controller} & \textbf{Fuel/veh} & \textbf{CO$_2$/veh} & \textbf{NO$_x$/veh} & \textbf{Stops/veh} \\
\midrule
Fixed-time & 51331.16 [50950.11, 51702.34] & 160933.10 [159738.47, 162096.82] & 63.527 [62.963, 64.078] & 0.61 [0.60, 0.62] \\
Queue-greedy & 48920.51 [48286.89, 49561.35] & 153375.72 [151389.26, 155384.84] & 59.999 [59.077, 60.935] & 0.78 [0.75, 0.81] \\
Max-pressure & 48945.34 [48296.34, 49598.18] & 153453.56 [151418.87, 155500.29] & 60.033 [59.090, 60.986] & 0.78 [0.75, 0.81] \\
DQN-RL & 50979.79 [50398.03, 51552.69] & 159831.68 [158007.78, 161627.78] & 62.980 [62.130, 63.817] & 0.79 [0.77, 0.82] \\
UCATSC-det & 47680.87 [47149.80, 48244.99] & 149489.30 [147824.32, 151257.90] & 58.213 [57.436, 59.033] & 0.69 [0.67, 0.72] \\
UCATSC-full & 47670.10 [47134.71, 48259.81] & 149455.53 [147777.03, 151304.37] & 58.198 [57.419, 59.059] & 0.69 [0.67, 0.72] \\
\bottomrule
\end{tabular*}
\end{table*}

\subsection{Scenario-Wise Delay and Constraint Stress Matrix}
\begin{figure}[!t]
    \centering
    \safeincludegraphics[width=\linewidth]{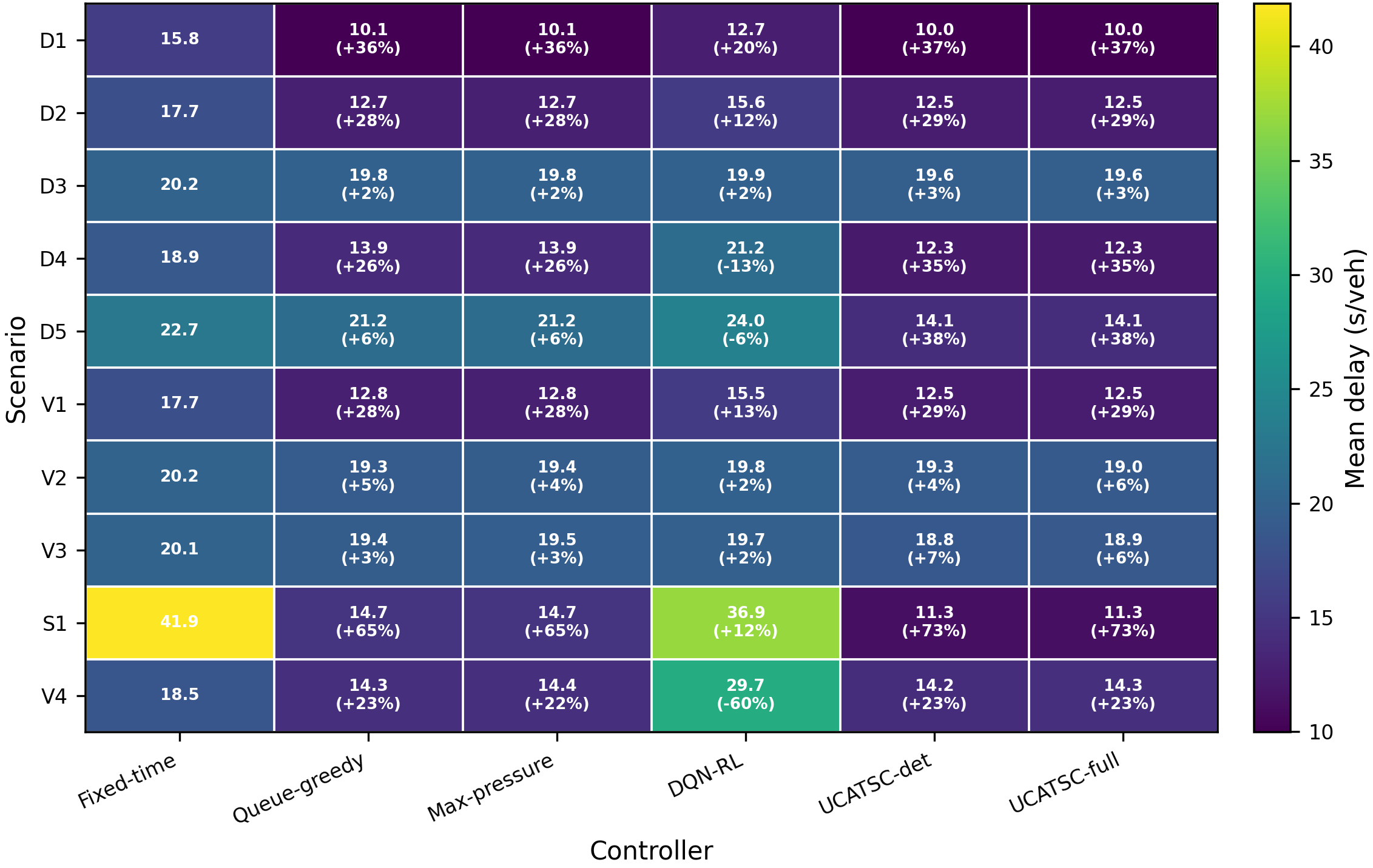}
    \caption{Scenario-wise mean delay heatmap including DQN-RL. Cell values denote mean delay in seconds per vehicle; secondary annotations show relative change versus fixed-time where included in the figure.}
    \label{fig:scenario_delay_heatmap}
\end{figure}

Table~\ref{tab:scenario_delay} and Fig.~\ref{fig:scenario_delay_heatmap} show scenario-wise delay across the primary and stress-test scenarios. The heatmap makes the aggregate result transparent: UCATSC-full is strongest in several high-demand and stress settings, whereas DQN-RL performs reasonably in some clean scenarios but degrades strongly in D5, S1, and V4.

\begin{table*}[!t]
\centering
\caption{Scenario-wise mean delay (s/veh) including the DQN-RL baseline.}
\label{tab:scenario_delay}
\scriptsize
\setlength{\tabcolsep}{4pt}
\begin{tabular}{lcccccc}
\toprule
\textbf{Scenario} & \textbf{Fixed-time} & \textbf{Queue-greedy} & \textbf{Max-pressure} & \textbf{DQN-RL} & \textbf{UCATSC-det} & \textbf{UCATSC-full} \\
\midrule
D1 & 15.82 & 10.06 & 10.06 & 12.70 & 9.99 & 9.99 \\
D2 & 17.72 & 12.75 & 12.75 & 15.62 & 12.52 & 12.52 \\
D3 & 20.17 & 19.83 & 19.83 & 19.86 & 19.60 & 19.60 \\
D4 & 18.85 & 13.89 & 13.89 & 21.24 & 12.31 & 12.31 \\
D5 & 22.70 & 21.24 & 21.24 & 24.04 & 14.12 & 14.12 \\
V1 & 17.74 & 12.77 & 12.79 & 15.47 & 12.55 & 12.54 \\
V2 & 20.19 & 19.26 & 19.43 & 19.83 & 19.29 & 19.04 \\
V3 & 20.12 & 19.42 & 19.49 & 19.73 & 18.76 & 18.90 \\
S1 & 41.87 & 14.74 & 14.74 & 36.90 & 11.30 & 11.30 \\
V4 & 18.55 & 14.28 & 14.38 & 29.74 & 14.20 & 14.33 \\
\bottomrule
\end{tabular}
\end{table*}

\begin{figure}[!t]
    \centering
    \safeincludegraphics[width=\linewidth]{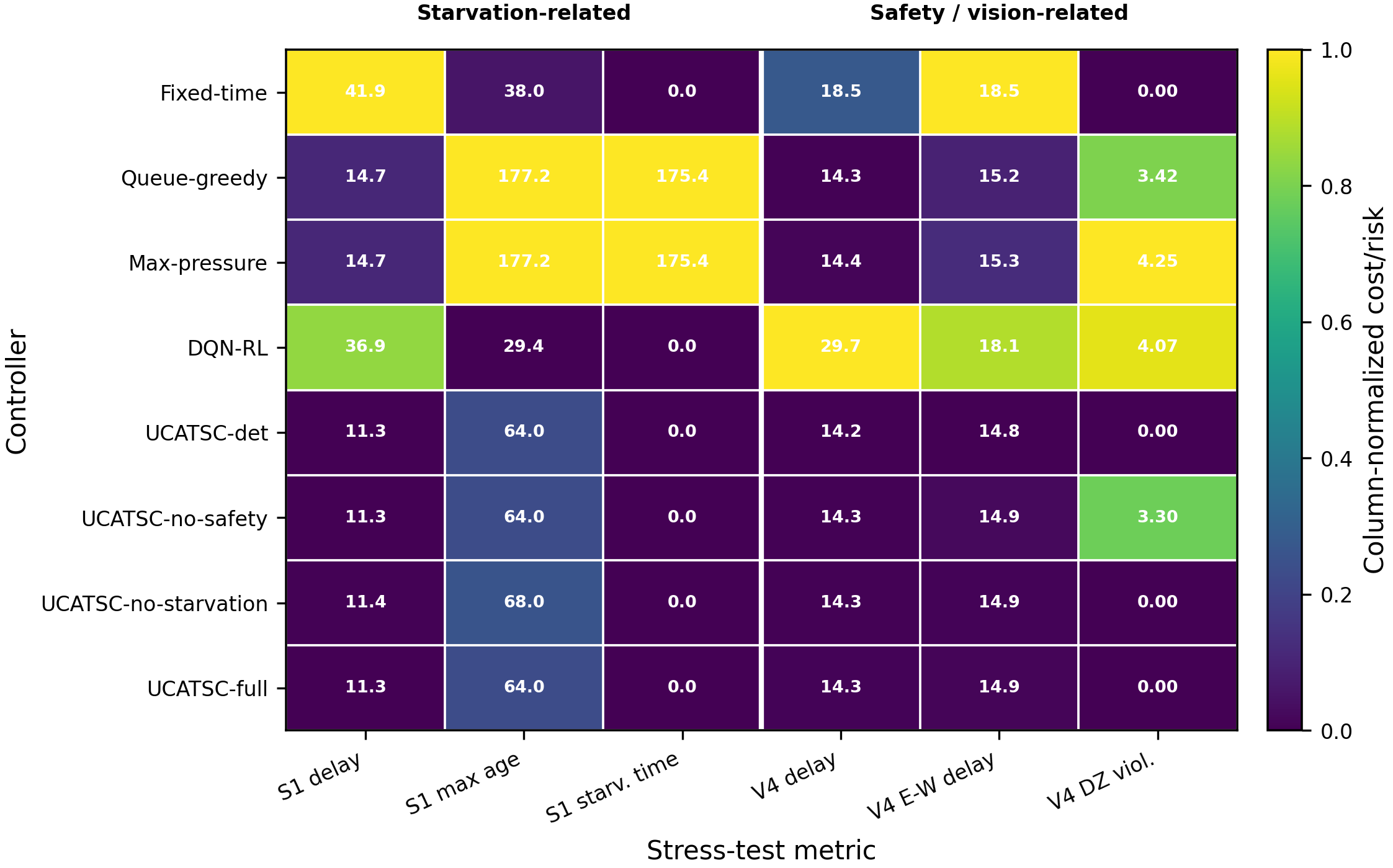}
    \caption{Constraint-focused stress-test metric matrix including DQN-RL. Columns are normalized independently for visualization, while overlaid numbers report the original metric values.}
    \label{fig:constraint_matrix}
\end{figure}

Figure~\ref{fig:constraint_matrix} summarizes the stress-test outcomes in a compact metric matrix. Each column is normalized independently so that lower values correspond to better outcomes within that metric; the overlaid numbers show the original metric values.

\subsection{Aggregate Performance Under Vision Degradation}
\begin{figure}[!t]
    \centering
    \safeincludegraphics[width=\linewidth]{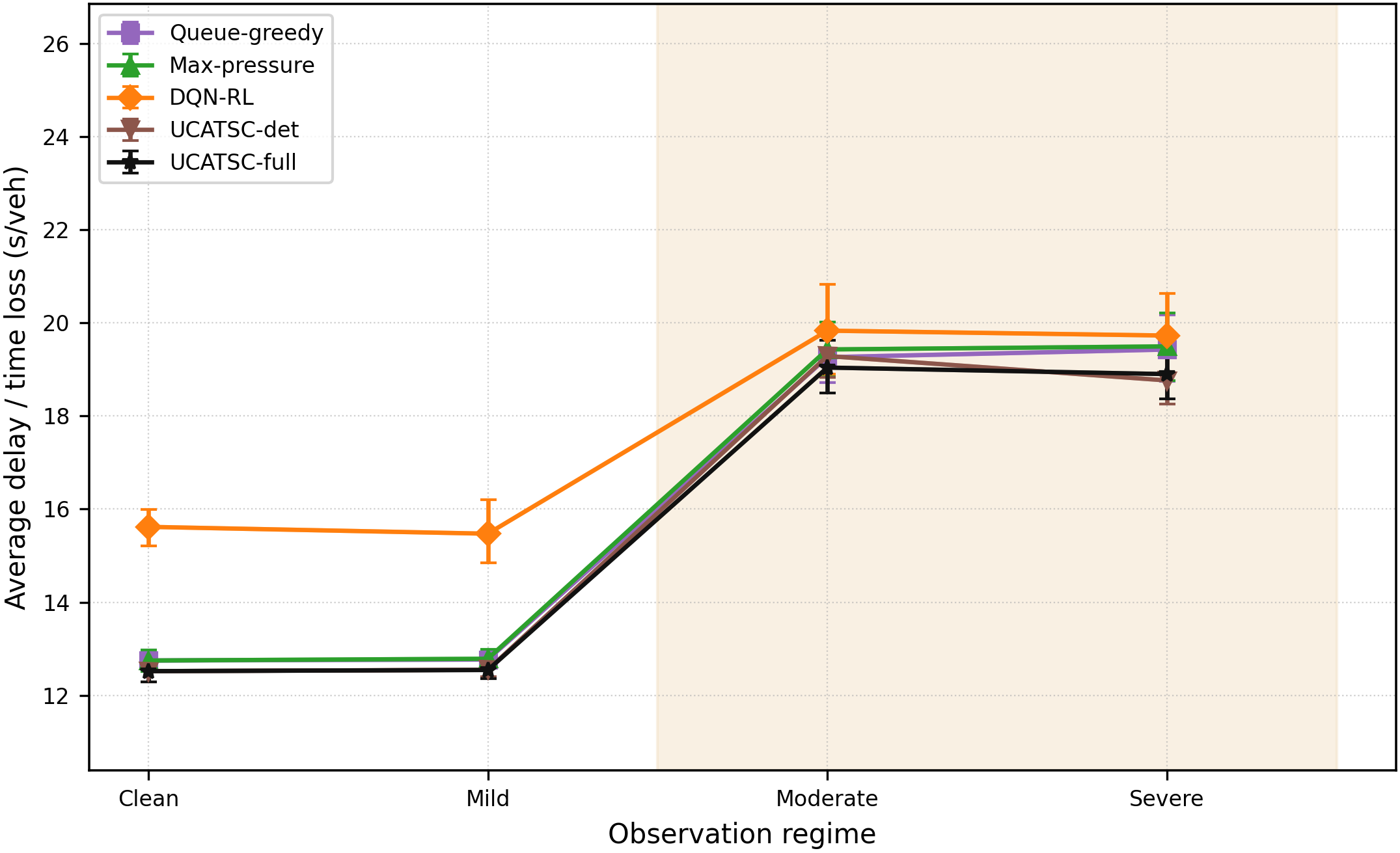}
    \caption{Effect of degraded observations on average delay across clean, mild, moderate, and severe sensing regimes, including the trained DQN-RL baseline.}
    \label{fig:vision_degradation}
\end{figure}

Figure~\ref{fig:vision_degradation} compares average delay across clean, mild, moderate, and severe sensing regimes. DQN-RL exhibits high variance under moderate degradation, while UCATSC-full and UCATSC-det remain close across the degradation regimes. The evidence does not support a claim that UCATSC-full strongly outperforms UCATSC-det in aggregate delay; instead, it supports robustness and safety-compatible operation under degraded observations.

\subsection{Reviewer-Safe Interpretation of the SUMO Evidence}
The main conclusion from the SUMO experiments is not universal delay optimality. The main conclusion is that explicit belief-space constraint checking changes the failure modes of adaptive signal control. Demand-responsive and learned policies can be competitive in delay, but they may initiate unsafe yellow transitions or under-serve minor movements in stress scenarios. Post-hoc safety masking can remove unsafe learned transitions, but in the targeted experiments it does not recover the integrated mobility and liveness behavior of UCATSC. UCATSC should therefore be interpreted as a safety- and service-aware constrained decision layer for uncertain vision-based signal control, not as a replacement for all adaptive or learned controllers in all networks.

\subsection{Runtime}
\begin{figure}[!t]
    \centering
    \safeincludegraphics[width=\linewidth]{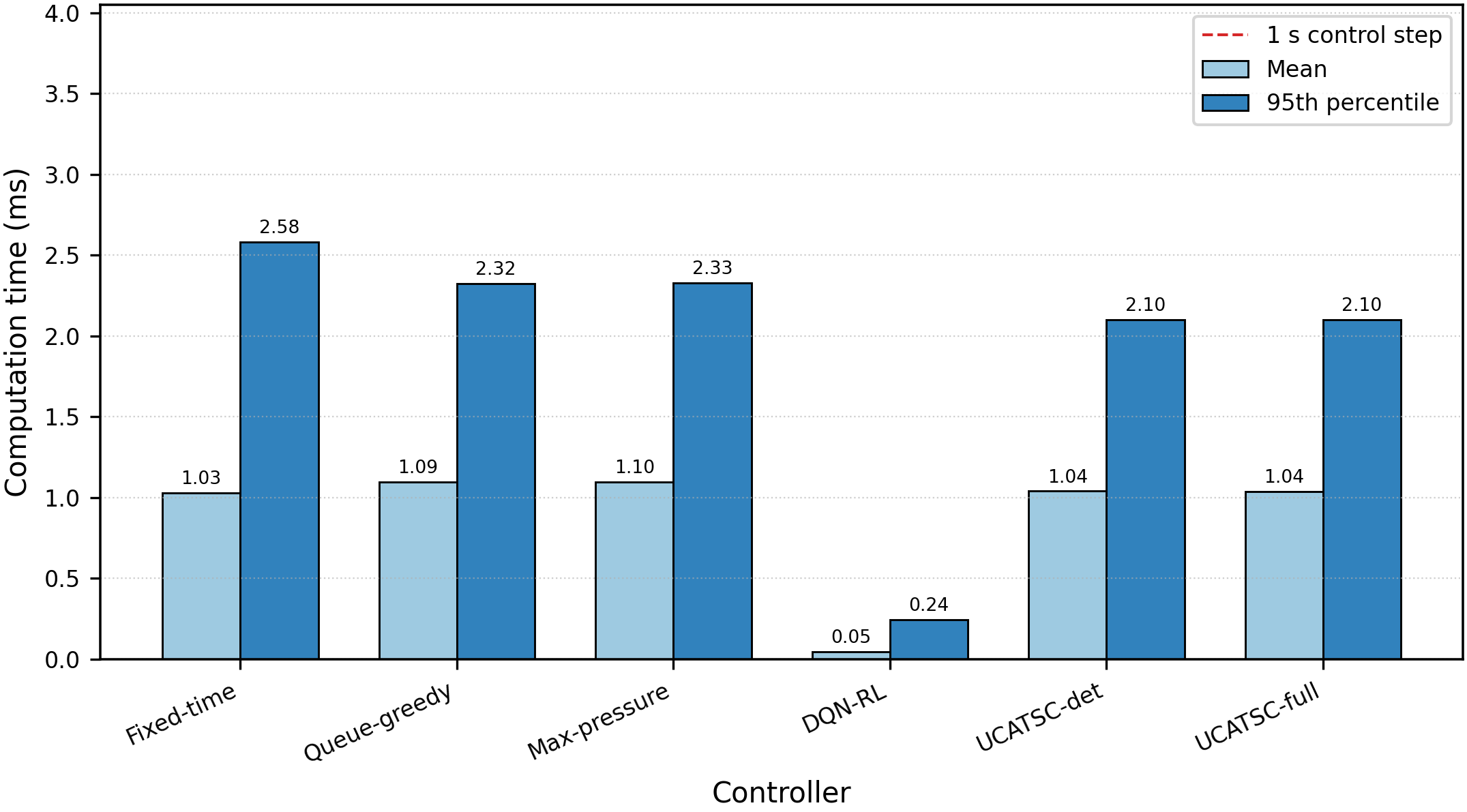}
    \caption{Runtime comparison across controllers including DQN-RL inference. Bars show mean computation time and 95th-percentile runtime; the red reference line indicates the one-second control step.}
    \label{fig:runtime}
\end{figure}

Table~\ref{tab:runtime} and Fig.~\ref{fig:runtime} report controller runtime. UCATSC-full has a mean computation time of 1.04 ms and a 95th-percentile runtime of 2.10 ms, which is well below the one-second control step used in the experiments. DQN-RL inference is faster, but it requires offline training and does not enforce the same predictive safety and service constraints.

\begin{table*}[!t]
\centering
\caption{Controller runtime in SUMO experiments. Values are mean [95\% bootstrap CI]. DQN-RL reports inference time only; training is performed offline.}
\label{tab:runtime}
\scriptsize
\setlength{\tabcolsep}{5pt}
\begin{tabular}{lccc}
\toprule
\textbf{Controller} & \textbf{Mean (ms)} & \textbf{95th percentile (ms)} & \textbf{Max (ms)} \\
\midrule
Fixed-time & 1.03 [0.97, 1.09] & 2.58 [2.44, 2.72] & 6.07 [5.56, 6.65] \\
Queue-greedy & 1.09 [1.02, 1.17] & 2.32 [2.18, 2.47] & 5.28 [4.81, 5.83] \\
Max-pressure & 1.10 [1.02, 1.17] & 2.33 [2.18, 2.47] & 6.20 [5.01, 8.01] \\
DQN-RL & 0.05 [0.04, 0.05] & 0.24 [0.21, 0.28] & 1.87 [1.74, 2.01] \\
UCATSC-det & 1.04 [0.97, 1.11] & 2.10 [1.98, 2.22] & 4.99 [4.48, 5.60] \\
UCATSC-full & 1.04 [0.97, 1.11] & 2.10 [1.98, 2.22] & 5.01 [4.49, 5.62] \\
\bottomrule
\end{tabular}
\end{table*}

\section{Three-Intersection Corridor Spillback Extension}
\label{sec:corridor_extension}

\subsection{Corridor Network and Evaluation Protocol}
To test whether the controller remains useful beyond one isolated junction, an additional SUMO experiment evaluates a three-intersection east--west arterial corridor with short inter-intersection links. The corridor contains three signalized intersections, denoted I1--I3, with two-phase operation at each junction. The east--west arterial links are intentionally short enough to create finite storage and potential spillback. Three demand scenarios are used: C1 is a balanced medium-demand corridor, C2 is a peak arterial spillback stress case, and C3 is a directional peak with stronger side-street pressure. Each scenario is evaluated with 20 matched seeds and 3600-s episodes.

The corridor comparison includes four controllers: fixed-time coordination, max-pressure, DQN-RL, and UCATSC-full. Fixed-time uses a coordinated cycle with arterial offsets. Max-pressure uses local upstream queue pressure with a downstream occupancy penalty. DQN-RL uses a shared independent policy at all three intersections with local queues, current phase, green age, and downstream occupancy summaries in the state. UCATSC-full applies the same predictive safety and service-age logic as the isolated-intersection controller and adds a downstream-storage term that discourages phase actions that would send vehicles into a nearly saturated inter-intersection link.

Spillback risk is measured using the maximum occupancy ratio across the four inter-intersection links. A spillback-threshold exceedance is counted whenever this maximum occupancy exceeds $0.85$. This metric is not a field-calibrated spillback probability; it is a reproducible stress indicator for finite-storage corridor behavior.

\subsection{Corridor Results}
Tables~\ref{tab:corridor_results} and~\ref{tab:corridor_emissions_runtime} summarize the corridor extension. UCATSC-full reduces mean delay from 68.80 s/veh under fixed-time control and 79.76 s/veh under max-pressure to 36.22 s/veh while preserving the same throughput level as fixed-time. It also reduces mean queue length from 24.30 vehicles under fixed-time and 17.05 vehicles under max-pressure to 9.37 vehicles. Per-arrived-vehicle CO$_2$ and NO$_x$ are also lowest for UCATSC-full. Max-pressure produces nonzero dilemma-zone violations in this corridor test, while UCATSC-full produces zero violations.

The DQN-RL corridor result must be interpreted jointly with throughput and service age. Although its delay over arrived vehicles is numerically low, it serves only 830.2 vehicles on average compared with approximately 2209 vehicles for fixed-time and UCATSC-full, and its maximum service age exceeds 5300 s. This indicates a learned under-service failure mode rather than a superior traffic-control policy. For this reason, the corridor DQN-RL result is used as evidence that a trained local RL policy can fail liveness and throughput requirements under distribution shift and finite-storage coupling.

\begin{table*}[!t]
\centering
\caption{Three-intersection corridor traffic, safety, and spillback results. Values are mean [95\% bootstrap CI] across three corridor scenarios and 20 matched seeds.}
\label{tab:corridor_results}
\scriptsize
\setlength{\tabcolsep}{3.5pt}
\begin{adjustbox}{width=\textwidth}
\begin{tabular}{lccccccc}
\toprule
\textbf{Controller} & \textbf{Delay (s/veh)} & \textbf{Queue (veh)} & \textbf{Throughput (veh)} & \textbf{Stops/veh} & \textbf{DZ/1000} & \textbf{Max age (s)} & \textbf{Spillback time (s)} \\
\midrule
Fixed-time & 68.80 [63.72, 73.81] & 24.30 [22.30, 26.28] & 2209.4 [2130.2, 2286.5] & 1.535 [1.447, 1.621] & 0.00 [0.00, 0.00] & 47.00 [47.00, 47.00] & 0.00 [0.00, 0.00] \\
Max-pressure & 79.76 [71.57, 87.88] & 17.05 [15.43, 18.71] & 2146.6 [2076.3, 2213.9] & 4.260 [3.832, 4.692] & 0.82 [0.35, 1.43] & 34.92 [32.85, 37.08] & 4.63 [1.93, 7.69] \\
DQN-RL & 2.56 [2.17, 3.00] & 49.88 [49.19, 50.56] & 830.2 [807.9, 851.3] & 0.149 [0.123, 0.178] & 0.00 [0.00, 0.00] & 5380.67 [5377.40, 5383.93] & 0.00 [0.00, 0.00] \\
UCATSC-full & 36.22 [33.27, 38.99] & 9.37 [8.48, 10.20] & 2209.8 [2130.6, 2286.6] & 1.758 [1.622, 1.889] & 0.00 [0.00, 0.00] & 74.73 [72.95, 76.47] & 48.68 [37.36, 60.10] \\
\bottomrule
\end{tabular}
\end{adjustbox}
\end{table*}

\begin{table*}[!t]
\centering
\caption{Three-intersection corridor SUMO-modeled per-vehicle emission outputs and runtime. Values are mean [95\% bootstrap CI]. DQN-RL reports inference time only.}
\label{tab:corridor_emissions_runtime}
\scriptsize
\renewcommand{\arraystretch}{1.08}
\setlength{\tabcolsep}{3pt}
\begin{tabular*}{\textwidth}{@{\extracolsep{\fill}}lccccc@{}}
\toprule
\textbf{Controller} & \textbf{Fuel/veh (mg)} & \textbf{CO$_2$/veh (g)} & \textbf{NO$_x$/veh (mg)} & \textbf{Mean runtime (ms)} & \textbf{95th perc. runtime (ms)} \\
\midrule
Fixed-time   & 92577.8 [87777.5, 97380.9]   & 290.24 [275.19, 305.30] & 124.53 [117.65, 131.38] & 0.000 [0.000, 0.000] & 0.000 [0.000, 0.000] \\
Max-pressure & 101399.4 [94494.8, 108343.7] & 317.90 [296.25, 339.67] & 137.06 [127.17, 146.99] & 0.221 [0.220, 0.223] & 0.241 [0.239, 0.244] \\
DQN-RL       & 292703.5 [285455.0, 300558.0] & 917.64 [894.92, 942.26] & 415.61 [405.19, 426.90] & 0.229 [0.227, 0.230] & 0.250 [0.247, 0.255] \\
UCATSC-full  & 64641.2 [61895.9, 67232.8]   & 202.66 [194.05, 210.79] & 84.23 [80.30, 87.95]    & 1.132 [1.055, 1.203] & 2.283 [2.132, 2.423] \\
\bottomrule
\end{tabular*}
\end{table*}

\begin{figure}[!t]
    \centering
    \safeincludegraphics[width=\linewidth]{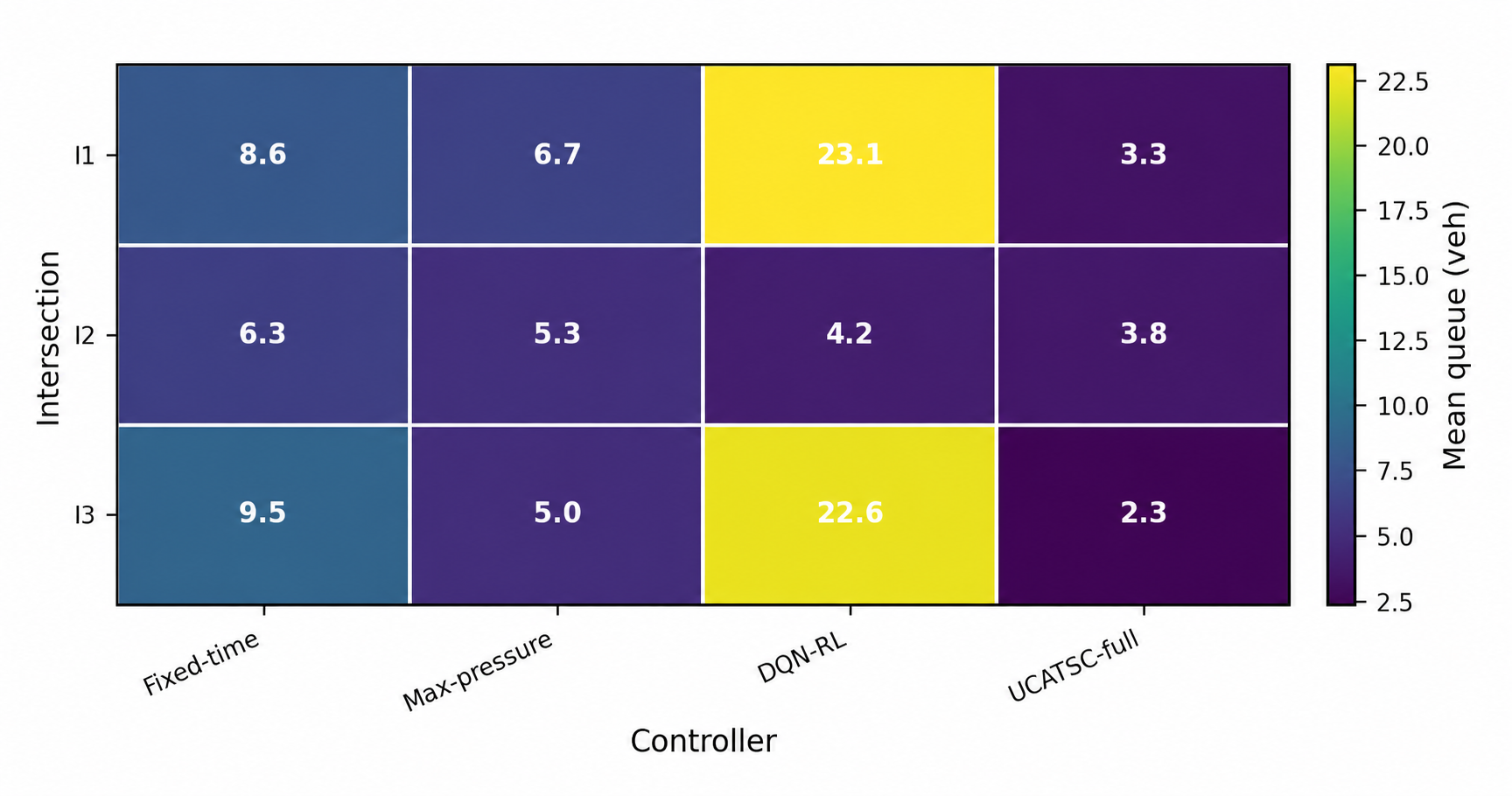}
    \caption{Intersection-wise mean queue heatmap in the three-intersection corridor extension. Cell values report mean queue length at I1, I2, and I3 for each controller.}
    \label{fig:corridor_queue_heatmap}
\end{figure}

\begin{table}[!t]
\centering
\caption{Scenario-wise corridor delay (s/veh).}
\label{tab:corridor_scenario_delay}
\footnotesize
\renewcommand{\arraystretch}{1.08}
\setlength{\tabcolsep}{3pt}
\begin{tabular*}{\columnwidth}{@{\extracolsep{\fill}}lcccc@{}}
\toprule
\textbf{Scenario} & \textbf{Fixed} & \textbf{MaxP} & \textbf{DQN} & \textbf{UCATSC} \\
\midrule
C1 & 42.09 & 42.69 & 2.06 & 20.82 \\
C2 & 88.29 & 118.82 & 2.75 & 39.14 \\
C3 & 76.01 & 77.76 & 2.88 & 48.69 \\
\bottomrule
\end{tabular*}
\end{table}

\begin{figure}[!t]
    \centering
    \safeincludegraphics[width=\linewidth]{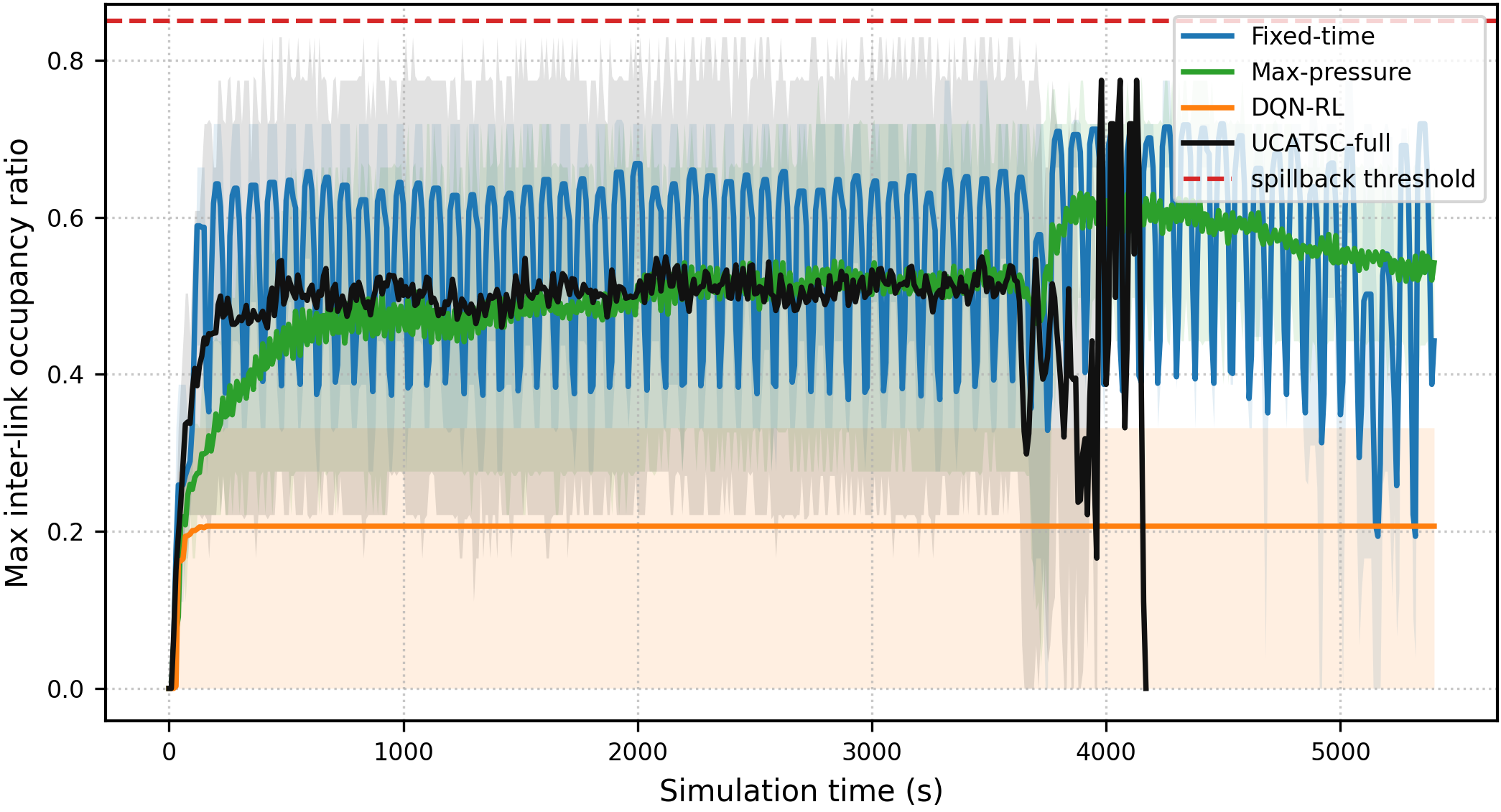}
    \caption{Corridor spillback-risk time series. The plotted signal is the maximum occupancy ratio across inter-intersection links; the dashed line marks the finite-storage stress threshold of 0.85.}
    \label{fig:corridor_spillback_ts}
\end{figure}

\begin{figure}[!t]
    \centering
    \safeincludegraphics[width=\linewidth]{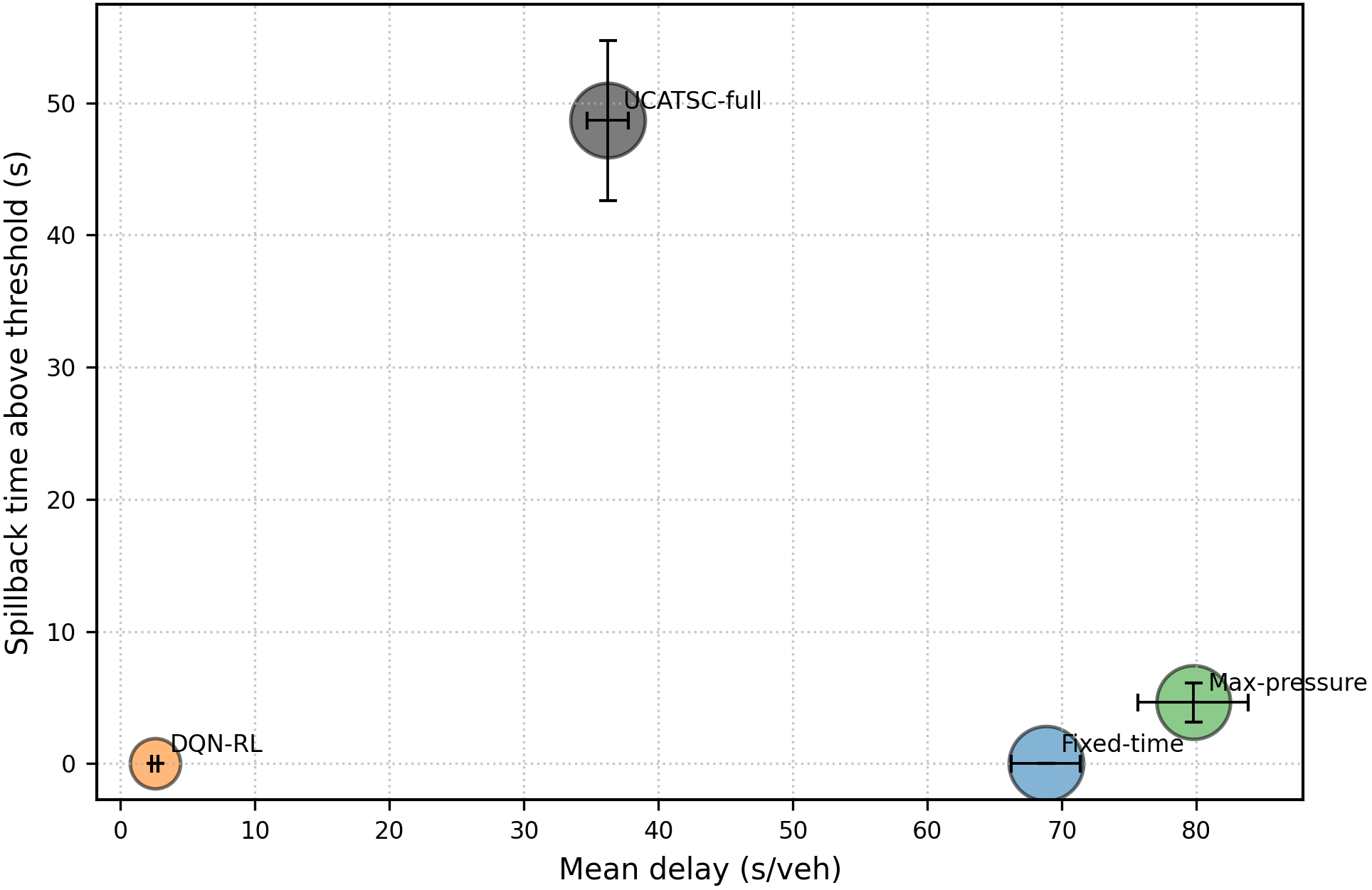}
    \caption{Delay--spillback tradeoff in the corridor extension. Marker size represents throughput, so the low-delay DQN-RL point should be interpreted together with its low served-vehicle count and high service age in Table~\ref{tab:corridor_results}.}
    \label{fig:corridor_pareto}
\end{figure}

\subsection{Corridor Interpretation}
Although UCATSC-full substantially reduces delay, queue length, and SUMO-modeled per-vehicle emission outputs, it records higher spillback-threshold exceedance time than fixed-time and max-pressure in the short-link corridor stress case. This indicates that the current local downstream-storage penalty improves mobility and queue burden but is not sufficient to fully suppress finite-storage spillback. Therefore, the corridor result should be interpreted as a multi-intersection stress extension rather than a complete network-level spillback solution. Larger arterial networks will require explicit offset optimization, stronger downstream storage prediction, and coordinated multi-intersection rollout objectives.

\section{Physical Vision-Testbed Supplement}
\label{sec:physical_testbed}

\subsection{Physical Vision-Based Testbed}
The supplementary physical evaluation uses a controlled small-scale intersection model with vision-based sensing. The dataset contains 8.75 hours of video from a fixed top-down camera at 1080p and 30 frames per second. Three crossing configurations were monitored under demand levels ranging from low to high density. The experiments include lighting changes and planned obstructions to emulate common perception-degradation conditions in camera-based intersection monitoring.

Vehicle detections and tracks are aggregated into three zone types: Queue Zones (QZ), Stop-Line Zones (SLZ), and a shared Conflict Zone (CZ). The controller does not use full raw images directly. Instead, it receives zone-level observations, detection confidence, and motion-state summaries.

\begin{figure}[!t]
    \centering
    \safeincludegraphics[width=0.95\linewidth]{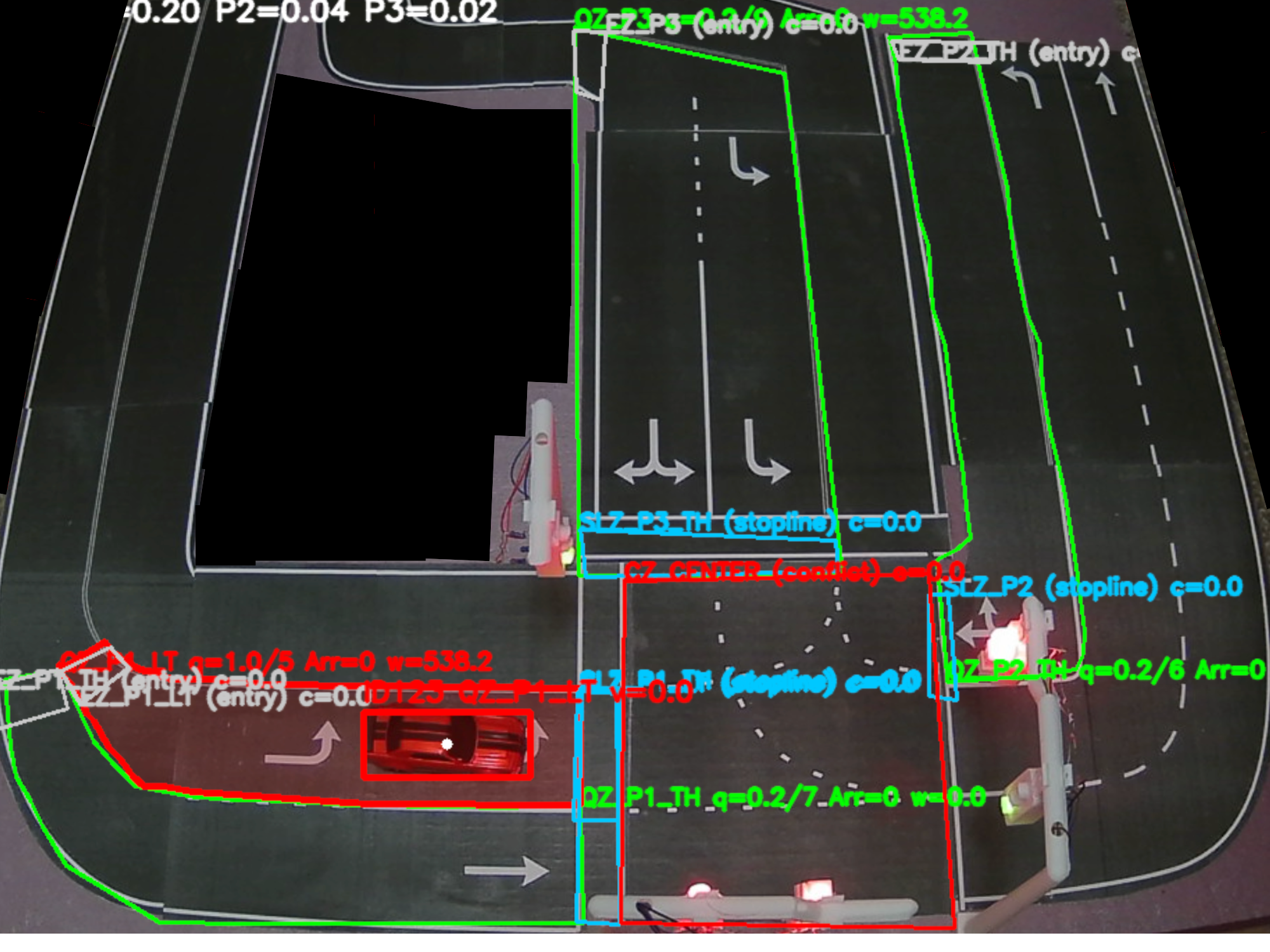}
    \caption{Vision-based perception and zone-level state abstraction in the physical testbed. Vehicle detections are aggregated into Queue Zones, Stop-Line Zones, and a Conflict Zone before being converted into movement-level belief variables.}
    \label{fig:testbed}
\end{figure}

\subsection{Vision-Testbed Proxy Metrics}
The physical testbed reports queue proxy, stopped-vehicle proxy, risk proxy, occlusion proxy, and runtime. These measures are useful for assessing the vision-to-control pipeline, but they are not direct measurements of delay, field emissions, or crash risk.

\begin{table}[!t]
\centering
\caption{Mean occlusion proxy $(1-p_{\mathrm{det}})$ with 95\% confidence intervals in the physical testbed. Lower values indicate better detection confidence.}
\label{tab:ci_occlusion}
\footnotesize
\renewcommand{\arraystretch}{1.08}
\setlength{\tabcolsep}{3pt}
\begin{tabularx}{\columnwidth}{@{}>{\raggedright\arraybackslash}X
                                >{\centering\arraybackslash}p{0.18\columnwidth}
                                >{\centering\arraybackslash}p{0.42\columnwidth}@{}}
\toprule
\textbf{Controller} &
\makecell[c]{\textbf{Relative}\\\textbf{change}} &
\makecell[c]{\textbf{Occlusion Proxy}\\\textbf{Mean [95\% CI]}} \\
\midrule
Queue-proxy &
\textcolor{ForestGreen}{$+0.0\%$} &
$0.03545\;[0.03476,\;0.03614]$ \\

UCATSC &
\textcolor{ForestGreen}{$+3.0\%$} &
$0.03438\;[0.03092,\;0.03783]$ \\

No EMA &
\textcolor{ForestGreen}{$+0.1\%$} &
$0.03540\;[0.03451,\;0.03629]$ \\

No hold &
\textcolor{ForestGreen}{$+5.8\%$} &
$0.03339\;[0.02875,\;0.03803]$ \\

No motion &
\textcolor{ForestGreen}{$+6.0\%$} &
$0.03332\;[0.03028,\;0.03637]$ \\
\bottomrule
\end{tabularx}
\end{table}

\begin{figure}[!t]
    \centering
    \safeincludegraphics[width=\linewidth]{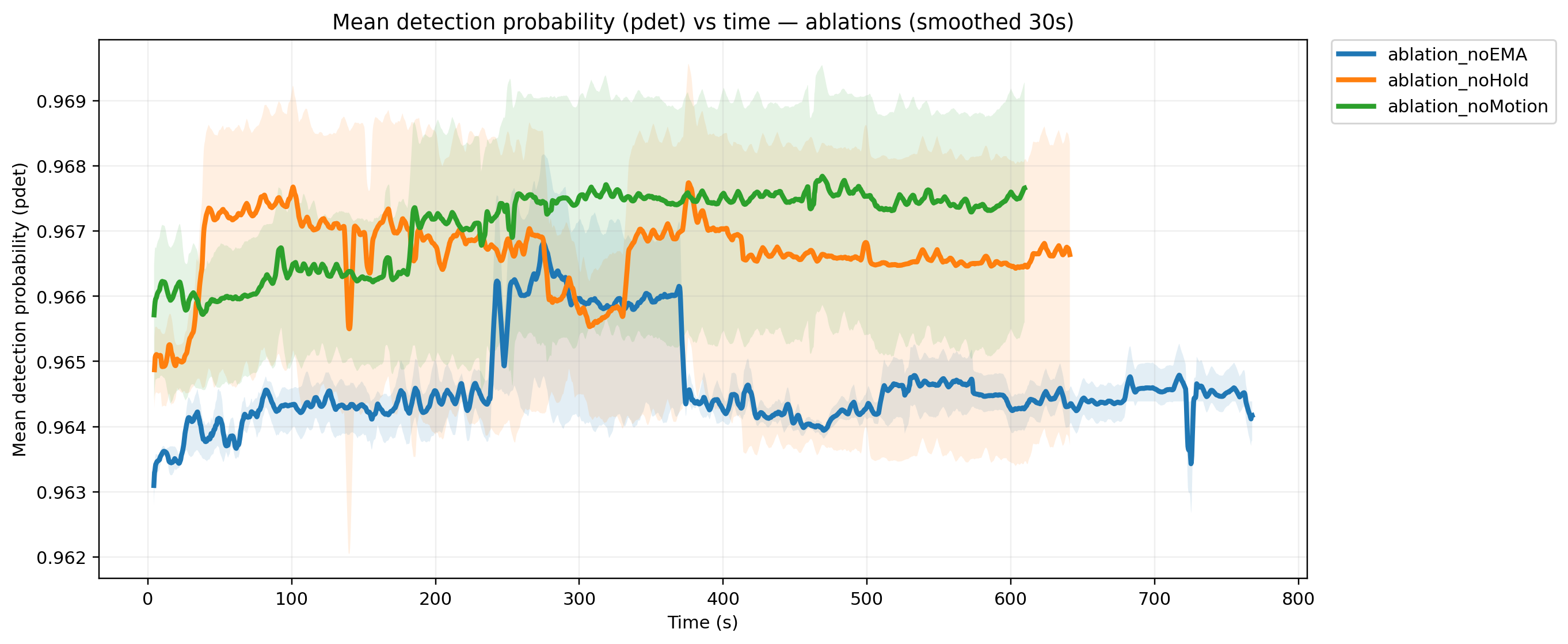}
    \caption{Mean detection probability $p_{\mathrm{det}}$ over time in physical vision-degradation scenarios.}
    \label{fig:pdet_ts}
\end{figure}

\begin{figure}[!t]
    \centering
    \safeincludegraphics[width=\linewidth]{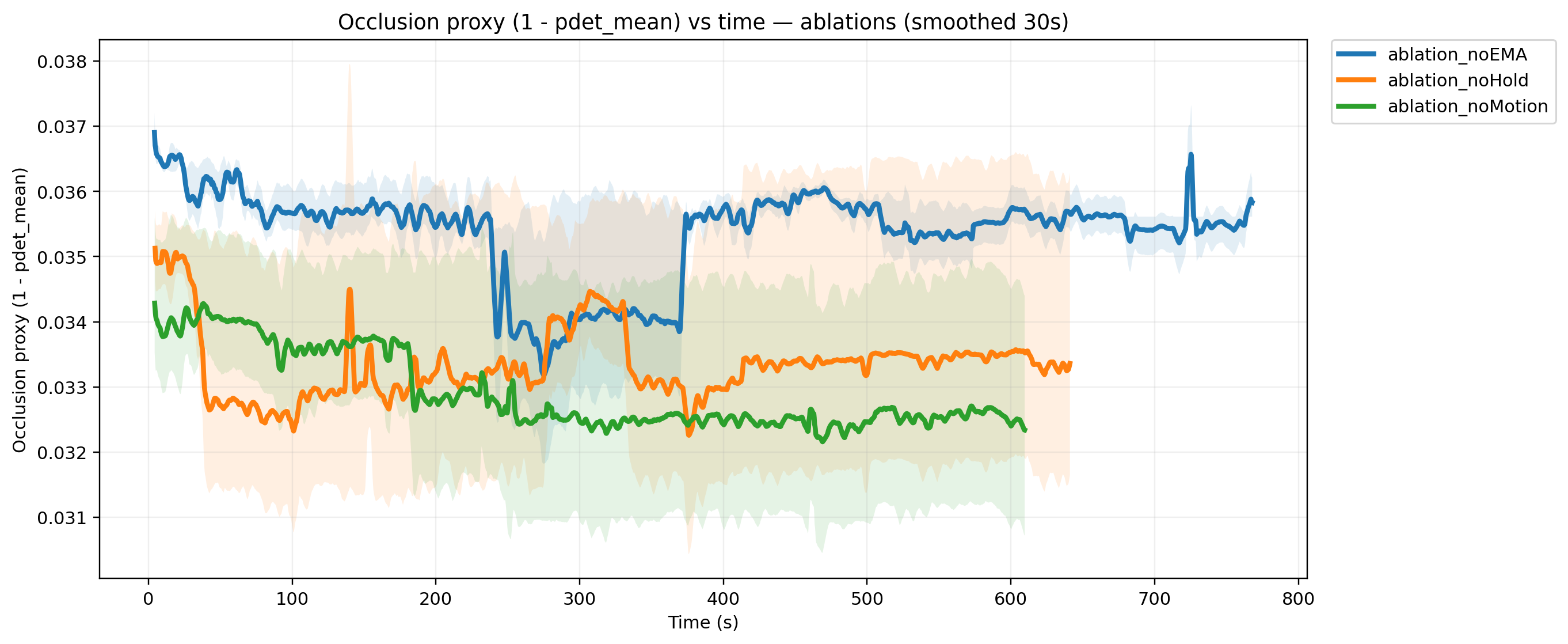}
    \caption{Occlusion proxy $(1-p_{\mathrm{det}})$ over time in physical vision-degradation scenarios. Lower values correspond to lower estimated visual uncertainty.}
    \label{fig:occ_ts}
\end{figure}

\begin{figure}[!t]
    \centering
    \safeincludegraphics[width=\linewidth]{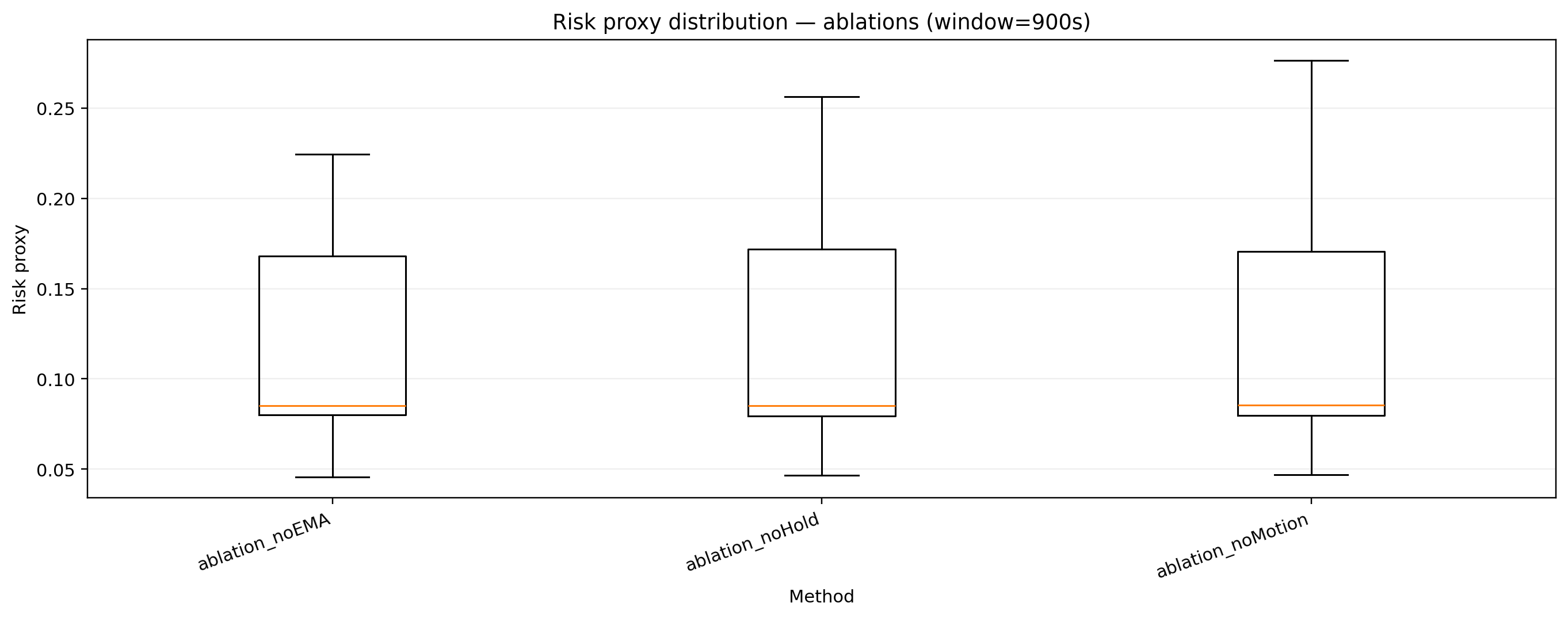}
    \caption{Risk-proxy distribution across physical-testbed controllers. The metric is a zone-derived indicator and is not interpreted as a direct crash-risk measure.}
    \label{fig:risk_box}
\end{figure}

\begin{table*}[!t]
\centering
\caption{Descriptive emission-linked proxy means in the physical testbed. These are not calibrated fuel or pollutant measurements and should be interpreted as exploratory proxy indicators.}
\label{tab:proxy_means}
\scriptsize
\setlength{\tabcolsep}{15pt}
\begin{tabular}{lcccc}
\toprule
\textbf{Controller} & \textbf{Idle Proxy} & \textbf{Queue Proxy} & \textbf{Risk Proxy} & \textbf{Total Proxy} \\
\midrule
Queue-proxy & 226.31 & 313.85 & 18.77 & 558.93 \\
Fixed-time & 110.27 & 174.68 & 9.57 & 294.53 \\
Occupancy-based & 131.00 & 185.53 & 10.78 & 327.30 \\
UCATSC & 118.45 & 187.61 & 10.58 & 316.63 \\
\bottomrule
\end{tabular}
\end{table*}

\begin{figure}[!t]
    \centering
    \safeincludegraphics[width=\linewidth]{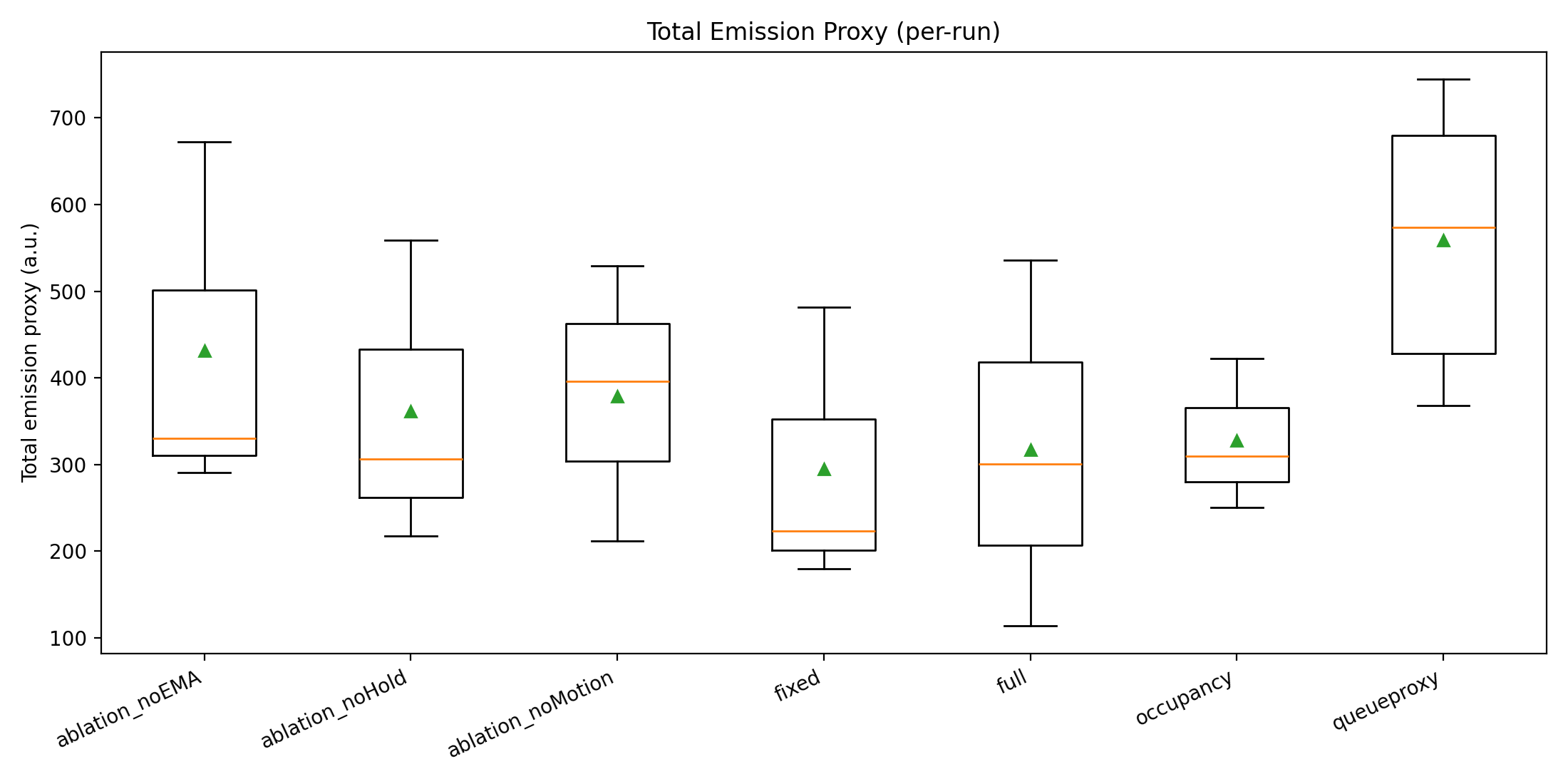}
    \caption{Total emission-linked proxy distribution across physical-testbed runs. This proxy compares stop--go behavior within the testbed but is not a direct emissions measurement.}
    \label{fig:proxy_box}
\end{figure}

\begin{figure}[!t]
    \centering
    \safeincludegraphics[width=\linewidth]{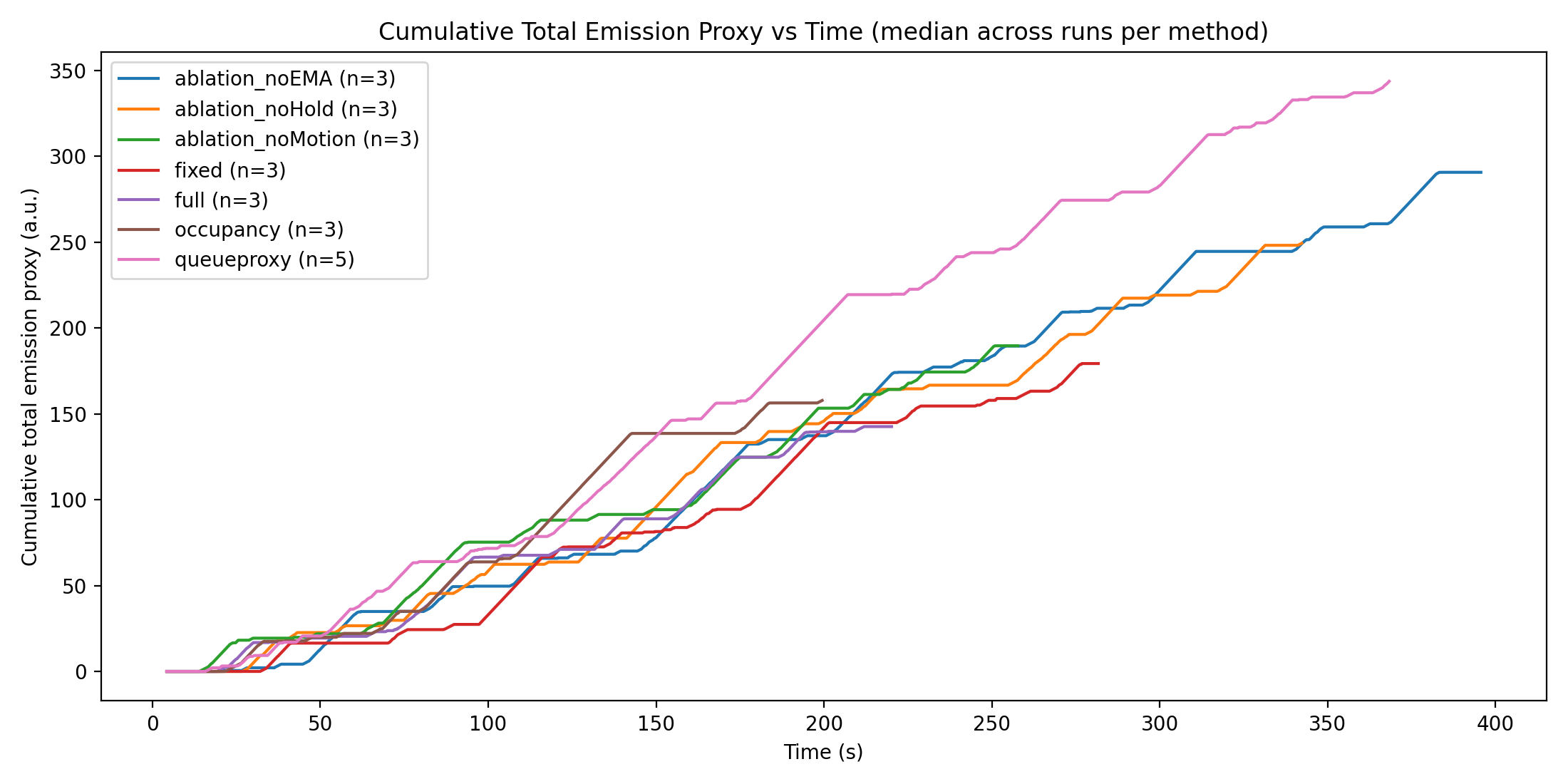}
    \caption{Median cumulative total emission-linked proxy over time in the physical testbed.}
    \label{fig:proxy_ts}
\end{figure}


\section{Discussion}

\subsection{Interpretation of the SUMO Results}
The strongest conclusions are drawn from SUMO because it provides matched seeds, ground-truth traffic states, standard traffic metrics, and reproducible controller comparisons. The results should be read as mechanism evidence for constrained action checking. In the primary isolated-intersection scenarios, UCATSC-full is competitive in delay and queueing and produces zero observed dilemma-zone violations. In S1 and S1X, the service-age mechanism prevents the minor movement from being indefinitely under-served. In V5, belief correction improves the under-detected E--W movement while preserving zero observed violations. In the corridor extension, UCATSC-full remains useful under finite-storage coupling, but local spillback handling does not yet constitute a complete network-level coordination solution.

\subsection{Safety and Liveness Interpretation}
The dilemma-zone results show that explicit predictive filtering removes unsafe yellow-onset decisions observed in queue-driven, max-pressure, DQN-RL, and safety-ablated policies. The starvation stress results show that queue-greedy and max-pressure can exceed the service-age limit under dominant-flow demand, while UCATSC-full keeps the minor-movement service age below the threshold. The targeted S1X ablation strengthens this conclusion by showing severe service-age growth when liveness safeguards are deliberately removed. DQN-RL avoids starvation in S1 but at the cost of high delay, and the safety-masked DQN eliminates violations only by producing a much less efficient policy in the targeted scenarios. These results show that satisfying one constraint by masking an action is not equivalent to solving the constrained traffic-control problem over a rollout horizon.

\subsection{Uncertainty-Specific Interpretation}
The strongest uncertainty-specific result is not the aggregate V1--V3 degradation curve, where UCATSC-det and UCATSC-full remain close. The stronger evidence comes from V5, where the error is movement-dependent and bursty. In that case, UCATSC-full allocates more balanced service to the under-detected E--W movement, reducing E--W delay and maximum E--W service age relative to UCATSC-det while keeping zero dilemma-zone violations. This supports a narrower but more credible claim: uncertainty-aware belief is valuable when perception errors are structured enough to bias service decisions, not as a universal guarantee of lower aggregate delay.

\subsection{Role of the Physical Testbed}
The physical vision testbed provides supplementary feasibility evidence for the sensing and implementation pipeline. It shows that zone-level perception, confidence-derived uncertainty, and online action evaluation can be integrated in a real camera-based prototype. It should not be used as the sole basis for traffic-performance or emissions claims.

\subsection{Interpretability of the Control Logic}
A practical advantage of UCATSC is that each action can be explained using the same variables used by the controller: expected queue burden, dilemma-zone risk, service-age constraints, and legal signal sequencing. This transparency may help diagnose failures caused by perception uncertainty, incorrect geometry, excessive conservatism, or service-age constraints.

\section{Limitations and Future Work}
\label{sec:limitations}

This study has several limitations. First, the main validation is performed in simulation, and simulation fidelity depends on route generation, car-following parameters, lane geometry, emission model settings, detector assumptions, and the reduced belief/risk model. Second, the physical evaluation is a controlled small-scale vision testbed rather than a city-scale field deployment. Its queue, stopped-vehicle, risk, and emission-linked measures are proxy indicators; they are not direct field measurements of delay, emissions, or crash risk. Third, the network experiment is a three-intersection corridor extension rather than a calibrated city-scale network with realistic offsets, turning movements, progression objectives, and multiple interacting intersections. Fourth, the RL comparison includes trained DQN-RL and safety-masked DQN-RL baselines, but it is not a complete benchmark against stronger methods such as PressLight, CoLight, FRAP/MPLight, or multi-agent RL. Fifth, the structured V5 scenario demonstrates a movement-specific benefit of belief correction under bursty under-detection, but it does not establish a universal aggregate-delay advantage of UCATSC-full over UCATSC-det in all degraded-vision regimes. Sixth, the conditional constraint properties are model-relative: they depend on sensing assumptions, threshold selection, feasibility of the constrained rollout problem, and accuracy of the risk estimator.

Future work should extend the corridor version to larger networks with coordinated progression, offset optimization, platoon modeling, and stronger spillback-aware queue dynamics. It should also integrate learned perception-uncertainty calibration from traffic-video datasets, benchmark stronger RL baselines such as PressLight, CoLight, FRAP/MPLight, and multi-agent RL under identical route seeds, and test the controller in hardware-in-the-loop or field-pilot settings.

\section{Conclusion}

This paper presented UCATSC, an interpretable uncertainty-aware constrained decision layer for vision-based adaptive signal control under partial observability. UCATSC maintains a reduced movement-level belief state, evaluates admissible phase actions through finite-horizon counterfactual rollouts, and filters actions using predictive dilemma-zone and service-age constraints. The evidence supports a scoped conclusion: in tested SUMO scenarios, UCATSC-full provides competitive mobility, eliminates observed dilemma-zone violations for constrained variants, bounds minor-movement service age in starvation stress tests, and operates with millisecond-level runtime. The mechanism-isolation experiments clarify each constraint. The S1X no-liveness ablation shows that removing liveness safeguards can produce severe E--W service-age growth, while UCATSC-full maintains zero starvation time. The V5 structured occlusion scenario shows that uncertainty-aware belief correction can improve the under-detected movement while preserving safety, although it does not guarantee lower aggregate delay in all degraded-vision regimes. The safety-masked DQN removes unsafe learned transitions but does not recover the integrated mobility/liveness behavior of constrained rollouts. The three-intersection corridor extension shows useful behavior under finite-storage coupling while revealing that local spillback handling is not yet a complete network-control solution. The physical vision testbed provides supplementary feasibility evidence for the zone-level vision-to-belief interface, not field validation of safety, emissions, or deployment readiness. Overall, UCATSC is best interpreted as a safety- and service-aware control layer for vision-based adaptive signal control when phase changes must be explainable under uncertainty. Its strength is integrating reduced belief-space counterfactual checking with explicit predictive safety and starvation-avoidance constraints.

\section*{Data and Code Availability}
The data and code used in this study are publicly released. Access link: \url{https://doi.org/10.5281/zenodo.19990731}.

\section*{Acknowledgment}
The authors gratefully acknowledge the open-source software used for computation, traffic simulation, computer vision, data processing, and visualization, including SUMO, NumPy, Pandas, SciPy, Matplotlib, OpenCV, and the Ultralytics YOLO framework.

ChatGPT-4 was used to refine the English language and improve grammar in this manuscript.

\bibliographystyle{IEEEtran}
\bibliography{references}

\end{document}